\documentclass[letterpaper, 10pt, conference]{ieeeconf}
\IEEEoverridecommandlockouts
\usepackage[utf8]{inputenc} % allow utf-8 input
\usepackage[T1]{fontenc}    % use 8-bit T1 fonts
\usepackage{hyperref}       % hyperlinks
\usepackage{url}            % simple URL typesetting
\usepackage{booktabs}       % professional-quality tables
\usepackage{amsfonts}       % blackboard math symbols
\usepackage{nicefrac}       % compact symbols for 1/2, etc.
\usepackage{microtype}      % microtypography
\usepackage{xcolor}         % colors
\usepackage{graphicx}
\usepackage{algorithm}
\usepackage{algpseudocode}
\usepackage{amsmath}
\usepackage{bm}
\usepackage{booktabs}
\usepackage{multirow}
\usepackage{graphics}
\usepackage{caption}
\usepackage{xspace}
\usepackage{wrapfig}
\newcommand{\Tref}[1]{Table~\ref{#1}}

\newcommand{\fref}[1]{Fig.~\ref{#1}}
\newcommand{\Fref}[1]{Figure~\ref{#1}}
\newcommand{\sref}[1]{Sec.~\ref{#1}}

\newcommand{\aref}[1]{Alg.~\ref{#1}}

\newcommand{\icra}[1]{#1}

\newcommand{\mb}[1]{\mathbf{#1}}
\newcommand{\modelfull}{Hypothesize, Simulate, Act, Update, and Repeat}
\newcommand{\modelname}{H-SAUR\xspace}

\newcommand{\argmax}{\mathop{\rm arg~max}\limits}
\newcommand{\etal}{\emph{et al.}\xspace}

\title{\LARGE \bf
    H-SAUR: Hypothesize, Simulate, Act, Update, and Repeat\\
    for Understanding Object Articulations from Interactions
}

\author{Kei Ota$^{1,2}$, Hsiao-Yu Tung$^{3}$, Kevin A. Smith$^{3}$, Anoop Cherian$^{4}$, \\Tim K. Marks$^{4}$, Alan Sullivan$^{4}$, Asako Kanezaki$^{2}$, and Joshua B. Tenenbaum$^{3}$% <-this % stops a space
	\thanks{$^{1}$Kei Ota is with Information Technology R\&D Center, Mitsubishi Electric Corporation, Japan.
		{\tt\small Ota.Kei@ds.MitsubishiElectric.co.jp}}%
	\thanks{$^{2}$Hsiao-Yu Tung, Kevin A. Smith, and Joshua B. Tenenbaum are Department of Brain and Cognitive Sciences, Massachusetts Institute of Technology, Cambridge, MA, USA.}%
	\thanks{$^{3}$Anoop Cherian, Tim K. Marks, and Alan Sullivan are with Mitsubishi Electric Research Labs, Cambridge, MA, USA.}%
}

\begin{document}
    \twocolumn[{%
    \renewcommand\twocolumn[1][]{#1}%
    \maketitle
    \begin{center}
        \centering
        \includegraphics[width=0.9\textwidth]{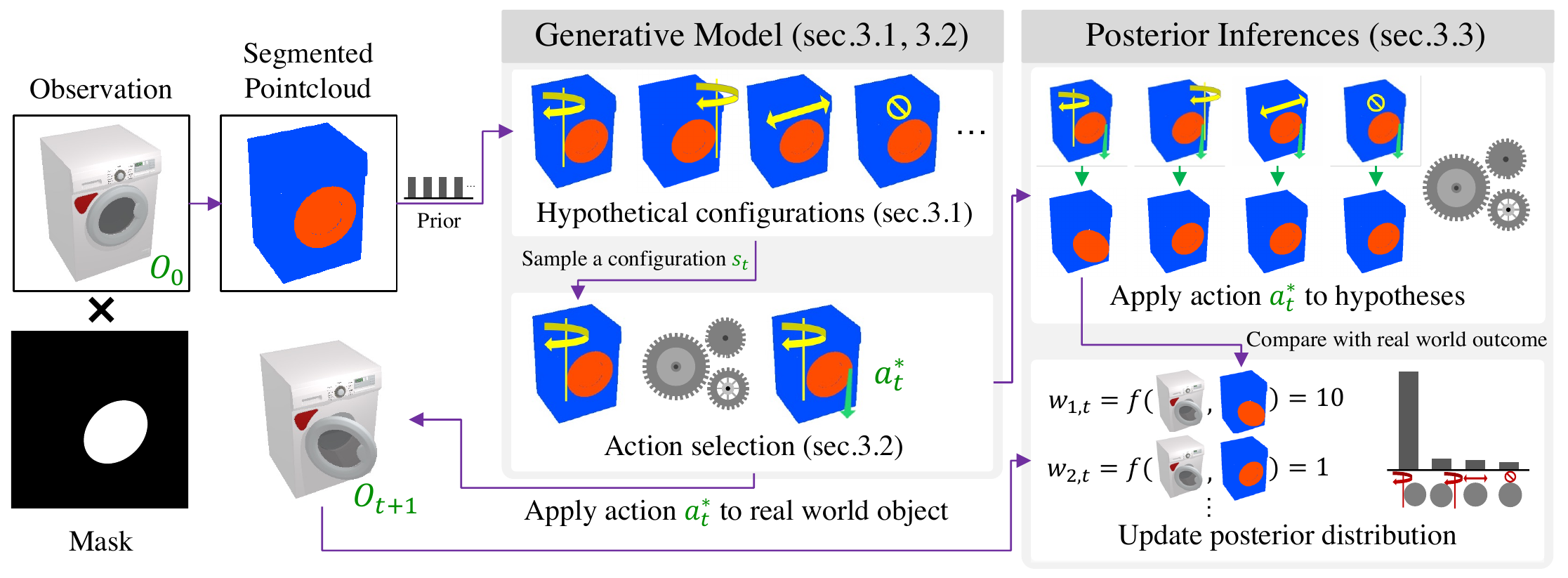}
        \captionof{figure}{
            \textbf{Overview of our ``{\color{red} Hypothesize}, {\color{orange} Simulate}, {\color{cyan} Act}, {\color{teal} Update}, and {\color{violet}Repeat}'' ({\color{red} H}-{\color{orange} S}{\color{cyan} A}{\color{teal} U}{\color{violet} R}) framework.} 
            \icra{We consider the task of estimating the kinematic structure of an unknown articulated object and use that structure for efficiently manipulating the object.}
            \textbf{Left:} A generative model produces several {\color{red} hypothetical} configurations given point cloud segments and {\color{orange} simulates} possible actions that maximally deform a sampled configuration. \textbf{Right:} By applying an {\color{cyan} action} and observing the outcome, the posterior inference is performed using the same generative model by simulating and {\color{teal} updating} the posterior distribution. We {\color{violet} repeat} the process until the convergence.
        }
        \label{fig:puzzle_box}
    \end{center}%
    }]
    \footnotetext[1]{Kei Ota is with Information Technology R\&D Center, Mitsubishi Electric Corporation, Japan.
        {\tt\small Ota.Kei@ds.MitsubishiElectric.co.jp}}%
    \footnotetext[2]{Kei Ota and Asako Kanezaki are with Tokyo Institute of Technology, Japan.}%
    \footnotetext[3]{Hsiao-Yu Tung, Kevin A. Smith, and Joshua B. Tenenbaum are with Department of Brain and Cognitive Sciences, Massachusetts Institute of Technology, Cambridge, MA, USA.}%
    \footnotetext[4]{Anoop Cherian, Tim K. Marks, and Alan Sullivan are with Mitsubishi Electric Research Labs, Cambridge, MA, USA.}%

    \thispagestyle{empty}
    \pagestyle{empty}
    \begin{abstract}
    %The world is filled with articulated objects that require particular actions to use, some of which are difficult or impossible to determine how to use from vision alone -- for instance, a door that might open inwards or outwards.
    %\AC{Articulated objects are ubiquitous in the real-world; however determining their articulations using vision alone is perhaps impossible, e.g., a door might open inwards or outwards.}
    The world is filled with articulated objects that are difficult to determine how to use from vision alone, e.g., a door might open inwards or outwards.
    Humans handle these objects with strategic trial-and-error: first pushing a door then pulling if that doesn't work. 
    %Robots that function in real world scenarios should also have the ability to learn about articulated objects from experience, and use this knowledge for future actions.
     %Humans and some other animals are shown to possess strong intrinsic motivation in 
     %exploring and researching objects that have unexpected and complex articulation mechanism. Though the process of strategic trial-and-error, most
     %people and animals can gradually grasp how the underlying mechanism works, and can use the knowledge to manipulate these objects at their will or in a creative way. 
     %In this work, we
     %try to replicate such ability on a machine. 
     We enable these capabilities in autonomous agents by proposing 
     % ``given hypotheses X: simulate,act,update,repeat'' (\modelname), 
     ``\modelfull'' (\modelname),
     %To produce these capabilities in a machine, we
     %propose "given hypotheses X: simulate,act,update,repeat" (\modelname), 
     a probabilistic generative framework that simultaneously generates a distribution of hypotheses about how objects articulate given input observations, captures certainty over hypotheses over time, and infer plausible actions for exploration and goal-conditioned manipulation. We compare our model with existing work in 
     %(1) estimating correct joint positions and types and (2)
     manipulating objects after a handful of exploration actions, on the \textit{PartNet-Mobility} dataset. 
     We further propose a novel \textit{PuzzleBoxes} benchmark that contains locked boxes that require multiple steps to solve.
     % To further evaluate how \modelname~handles objects that require multiple interactions, we design a novel benchmark, \textbf{PuzzleBoxes} that contains boxes that require a sequence of steps to solve. 
     We show that the proposed model significantly outperforms the current state-of-the-art articulated object manipulation framework, despite using zero training data. We further improve the test-time efficiency of \modelname by integrating a learned prior from  learning-based vision models.  
     
     %common sense including 3D scene understanding 
%for evaluating the performance of the proposed model and the baselines in (1) estimating correct joint positions and types and (2) manipulating the object after a handful of exploration actions. 
\end{abstract}
    \IEEEpeerreviewmaketitle
    \section{Introduction}

Every day we are surrounded by a number of articulated objects that require specific interactions to use: our laptops can be opened or shut, windows can be raised or lowered, and drawers can be pulled out or pushed back in. A robot designed to function in real-world contexts should thus be able to understand and interact with these articulated objects.

Recent advances in deep reinforcement learning (RL) have focused on this problem and enabled robots to manipulate articulated objects such as drawers and doors \cite{googlearm, doorgym, wu2022vatmart, xu2022umpnet}. However, these systems typically produce fixed actions based on observations of a scene, and thus, when the articulated joint is ambiguous (e.g., a door that slides or swings), they cannot adapt their policies in response to failed actions. While some systems attempt to adjust policies during test-time exploration to recover from failure modes \cite{adaafford, maml}, they only propose local action adjustments (pull harder or run faster) and so are insufficient in cases where dramatically different strategies need to be applied, e.g.,  from ``sliding the window'' to ``pushing the window outward from the bottom.''

In contrast, humans and many other animals can quickly figure out how to manipulate complex articulated man-made objects, e.g., puzzle boxes, with very little training \cite{Thorndike1911-THOAI, Auersperg2013ExplorativeLA, 10.1525/bio.2011.61.9.7}. These capabilities are thought to be supported by rapid, strategic trial-and-error learning -- interacting with objects in an intelligent way, but learning when actions lead to failures and updating mental representations of the world to reflect this information \cite{allen2020rapid}. We argue that robotic systems that can learn how to manipulate articulated objects should be designed using similar principles.

In this work, we propose ``\modelfull'' (\modelname), an exploration strategy that allows an agent to figure out the underlying articulation mechanism of man-made objects from a handful of actions. At the core of our model is a probabilistic generative model that generates hypotheses of how articulated objects might deform given an action. 
%Given a scene with a visually ambiguous kinematic object, 
Given a kinematic object, our model first generates several hypothetical articulation configurations of the object from 3D point clouds segmented by object parts. 
% 3D segments of the object parts obtained from the observed point cloud. 
Our model then evaluates the likelihood of each hypothesis through analysis-by-synthesis -- 
%given the 3D geometry of the object parts,
%and potential joint configurations, 
the proposed model simulates objects representative of each hypothetical configuration, using a physics engine to predict likely outcomes given an action. The virtual simulation helps resolve three critical components in this interactive perception setup: (1) deciding real-world exploratory actions that might produce meaningful outcomes, (2) reducing uncertainty over beliefs after observing the action-outcome pairs from real-world interactions, (3) generating actions that will lead to successful execution of a given task after fully figuring out the articulation mechanism.
%simulates physically plausible outcome of the object by constructing a virtual object segments and potential kinematic joints in a physics engine. 
% copy from Ota's slides
The contributions of this paper can be summarized as follows:
\begin{enumerate}[leftmargin=*,itemsep=0mm]
    \item We propose a novel exploration algorithm for efficient exploration and manipulation of puzzle boxes and articulated objects, by integrating the power of probabilistic generative models and forward simulation. Our model explicitly captures the uncertainty over articulation hypotheses.
    %possible articulation configurations over time.
    \item We compare \modelname against existing state-of-the-art methods, and show it outperforms them in operating unknown articulated object, despite requiring many fewer interactions with the object of interest. 
    \item We propose a new manipulation benchmark -- \textit{PuzzleBoxes} -- which consists of locked boxes that require multi-step sequential actions to unlock and open, in order to test the ability to explore and manipulate complex articulated objects.
    % To study efficient exploratory behavior during deployment time, w
    %of articulated objects with visually ambiguous mechanism that require test-time exploration to learn about the mechanism and eventually open it.
\end{enumerate}

%We proposed a human-inspired method to interact with articulated objects: 
%a combination of a particle filter approach to keep track of uncertainty in different articulation models and their parameters, and the use of a physics engine to predict the future given a hypothesis that works like a mental simulator.
%We demonstrated that our proposed approach leads to a sample-efficient estimation of joint configurations and affordance maps. 
%Our method solves the task within approximately ten interactions.
%We proposed a dataset, which consists of a set of complex manipulation tasks that have more than two joint chains, which are common in everyday life, but have not been explored so far.

    \section{Related Work}
% exploration policies
{\noindent \bf Kinematic Structure Estimation.}
% Knowing the articulation mechanism of a man-made object can facilitate efficient manipulation of the object.
% To efficiently manipulate an articulated object, a natural first step is to predict the articulation mechanism of the object.
A natural first step to manipulate an object is to predict the articulation mechanism of the object.
Li \etal \cite{articulated_li} and Wang \etal \cite{articulated_wang} proposed models to segment object point clouds into independently moving parts and articulated joints.
However, this requires part and articulation annotations, and thus does not generalize to unexpected articulation mechanisms.
% However, this requires part segments and articulation labels, and the trained models cannot handle unexpected articulation mechanisms.
% proposed models that take the point cloud of an object and segment it into independently moving parts and articulated joints connecting the parts. Yet, training these models requires large datasets with part segments and articulation labels, and the trained models cannot handle unexpected articulation mechanisms that deviate from their categorical priors. 
%Moreover, some articulated objects possess unexpected articulation mechanisms that deviate from their categorical priors. 
Previous work address this by proposing to visually parse articulated objects under motion \cite{ditto, SturmSB14, heppert2022categoryindependent, BagnelKatz-2013-7694, zhoushi, DBLP:journals/corr/CifuentesIWSB16}. % to leverage the strong visual cues from motion. 
Yet, most work assumes the objects are manually articulated by humans or scripted actions from the robot.
%, which are unavailable in a real world setup. 
In this paper, we study how an agent can jointly infer articulation mechanism and exploratory actions that helps to reveal the articulation of an object, i.e., in an interactive perception setup \cite{interactiveperception}. Niekum \etal \cite{niekum2015ActiveArtic} addresses a similar setup, but only handles articulated objects with a single joint
and assumes the robot knows where to apply forces.
%only needs to decide the direction to pull/push. 
%The work also needs VR tag to track the 6DoF pose of each object part. 
Kulick \etal \cite{lockbox} and Baum \etal \cite{baumlockbox} handle dependency joints but assume each joint is either locked or unlocked, which is  ambiguous for general kinematic objects. 
%require 6DoF tracking for object parts and a detector that classifies a joint as locked or unlocked, whose definition is ambiguous on general kinematic objects. 
\modelname  takes raw point clouds and part segmentations as inputs, and infers both the joint structure of the object and how to act. This model can handle articulated objects with an arbitrary number of joints and joint dependencies by leveraging off-the-shelf physics simulation for general physical constraint reasoning.

%cite works in RL that aims at efficient exploration

%Please check \cite{interactiveperception} for a more complete survey on interactive perception.

{\noindent \bf Model-free approaches for manipulating articulated objects.}
Instead of explicitly inferring the articulation mechanism, recent works in deep RL learn to generate plausible object manipulation actions from pointclouds \cite{mo2021where, wu2022vatmart, adaafford}, RGB-(D) images \cite{xu2022umpnet, googlearm, doorgym}, or the full 3D state of the objects and their segments \cite{metaworld, opal,pertsch2021skild}.
While most of these RL approaches learn through explicit rewards, %recent advancement of self-play imitation learning how to interact with the environment through reward-less exploration.
recent approaches have learned to manipulate objects in a self-supervised manner, through self-driven goals or imitation learning~\cite{lynch2019play2,dinyari2020learning}.
%$Learning from self-play frameworks also acquires skills to manipulate articulated objects~\cite{lynch2019play2,dinyari2020learning}.} 
However, all of these systems require a large number of interactions during training and cannot discover hidden mechanisms that are only revealed through test-time exploratory behaviors.
\icra{Furthermore, while they focus on \textit{training-time} exploration, our work focuses on \textit{testing-time} exploration where only a small number of interactions is permitted.}

\section{Method}
%Our approach is inspired from human structure reasoning when interacting with articulated objects: humans first estimate the object's structure, such as what kind of joint does the object have, and then generate an action based on the estimation. After observing the interactions with the object, humans can quickly update their own believes over the articulation mechanisms. 
%For example, when we see a closed door, we will have hypotheses of the structure of the object, such as the door is attached to the frame by revolute joint, prismatic joint, and so on. Then, we try to apply an action, such as pushing the door, and if it fails we can quickly update the hypotheses and try different action, such as pulling the door.
\icra{We consider a task of estimating kinematic structure of an unknown articulated object and use the estimation for efficient manipulation. We are particularly interested in manipulating a visually ambiguous object, e.g., a closed door that can be opened by pulling, pushing, sliding, etc. In such a situation, the agent needs to estimate its underlying kinematic configuration, and update its beliefs over different configurations based on the outcome of past failed actions.}
% The challenge here is the agent needs to deeply understand the underlying kinematic configuration, and update action distributions based on the past failed actions, e.g., the agent pulls a door if it failed to push the door before.}
% We are particularly interested in how a probabilistic framework can track uncertainty over underlying kinematic configurations and how we can efficiently resolve the ambiguity by using a physics engine that works as mental representations of humans.}

We propose ``\modelfull'' (\modelname), a physics-aware generative model that represents an articulated object manipulation scene in terms of 3D shapes of object parts, articulation joint types and positions of each part, actions to apply on the object, and the change to the object after applying the actions.
%In this section, we provide details of the proposed method that realizes the efficient object's structure reasoning.
%To do that, we make use of particle filters to track the transitions of the different hypotheses, and a physics engine to predict the future caused by interacting with an articulated object.
\icra{In this work, we assume to have access to a physics engine that can take as input 3D meshes (estimated from
a point cloud) of a target unknown object with an estimated kinematic configuration, and produce hypothetical simulated articulations of this object when kinematically acted upon.}
The method consists of three parts. 
First, we initiate a number of hypothetical configurations that imitate a target object
%from point cloud observations of a target object 
by sampling articulation structures from a prior distribution.
The prior distribution can be uniform or from learned vision models.
%First, from the point cloud observation of a target object, we initiate a number of hypothetical articulated objects whose articulation structures are sampled from a prior distribution.
%synthetic replicas of the target object using the generative model. The replicas are with different articulation structures sampled from a prior distribution.
%, and initialize the prior distribution over the different hypothetical structures. 
Second, we sample one of the hypotheses to generate an action that is expected to provide evidence for or against that hypothesis. Finally, we apply the optimal action to the target object and update beliefs about object joints based on the outcome.

\subsection{Generating Hypothetical Articulated Objects}
%We first introduce a probabilistic model on which we track probabilities of different hypotheses.
Given the observed pointcloud $O$ of a target object along with its part segmentation, $m,$ we generate a number of kinematic replicas of the object.
Since the true articulation mechanism is initially unknown, we generate these replicas by sampling different kinematic structures from uniform prior distributions over joint types and parameters.
% Since the true articulation mechanism is assumed to be unknown in the initial frames, we generate these replicas by assigning them  with different kinematic structures sampled from some prior distribution.

%Since the true articulation mechanism is unknown in the initial frames.

\noindent \textbf{Object Parts.}
From the observed pointcloud $O$ and segmentation masks, $m_1, m_2, \cdots, m_{N_v},$ where $N_v$ is number of available views, we can break the pointcloud into part-centric pointcloud $O^1, O^2, \cdots, O^{N_p}$ where $N_p$ is the total number of object parts.
%Given a pointcloud of the closed target link and a part segmentation mask, we generate a hypothetical object (template).
%First, we make hypotheses of an object's structure from an observation using a vision sensor. 
% Although humans generally estimate the structure in an implicit way by exploiting the past experiences, sometimes using a strong visual prior, here we do not assume we have such strong priors.
% Instead, we make hypotheses from observations of the test object.
%Given a pointcloud of the closed target link and a part segmentation mask, we generate a hypothetical object (template).

\noindent \textbf{Articulation Joints.}
Each object part is attached to a base of the object with a joint. We consider three most common types of articulation joints: revolute (r), prismatic (p), and fixed (f). For revolute and prismatic joints, we further generate possible joint axes and positions, using the tight bounding boxes fitted to the part-centric pointcloud to obtain a total of ${J}$ possible joints. 
% More details are in the supplementary. 
The $j^\text{th}$ joint is denoted as $\theta^{(j)} = (c, d)$ where $c\in \{r, p,f\}$ is the joint type and $d \in \mathbb{R}^6$ is the 6-DoF pose of the joint axis. The prior distribution $p(\theta^{(j)})$ for the joint type is assumed to be uniform at $t=0.$ One can also use learned prior from vision models that predict joint types.

%Aside from the joint types and axis, 
In addition,
most articulated joints have lower and upper limits of how much the joint can be deformed. We denote the limits as $\theta^\text{low}$ and  $\theta^\text{high}.$ The prior distribution is sampled uniformly from $[-\theta_\text{MAX}, 0]$ and $[0, \theta_\text{MAX}],$ respectively. The full state of the joint for object part $O^i$ is $s^i = (\theta^{(\sigma(i))}, \theta^{\text{low}_i}, \theta^{\text{high}_i}, \theta^{\text{cur}_i})$, where $\sigma(i) \in \{1, 2, \cdots, J\}$ is the joint configuration for the $i^\text{th}$ object part, and $\theta^{\text{cur}_i}$ is the joint position at the current time step. The prior over all the latent variables is:
\begin{equation}
\label{eq:prior}
    p(s^{1:N_p}) =  \prod_{i=1}^{N_p} p(\theta^{(\sigma(i))}) p_{\text{unif}[-\theta_\text{MAX}, 0]}(\theta^{\text{low}_i}) p_{\text{unif}[0, \theta_\text{MAX}]}(\theta^{\text{high}_i}).
\end{equation}
We approximate the distribution by maintaining a particle pool, $\mathcal{S},$ where each particle in the pool represents a particular setup for the articulation configurations.

%We denote a joint with $j = (c, d)$

\subsection{Simulating and Selecting Informative Action} \label{sec:action}
%Given geometry shapes of object parts and the joints connecting them, one can use any off-the-shelf physics engine to simulate the outcome given an action. In this work, we use SAPIEN \cite{xiang2020sapien} for the physics simulation. The physics simulation serves two critical roles: (1) Given known articulation, we can use the simulation to figure actions that might lead to 
%informative deformation; (2) Given unknown articulation, we can use the simulation to examine which hypotheses match with the observation in the real world. Here we focus on the first point to select informative action.

We utilize virtual simulations to generate an optimal action that reduces the uncertainty of joint configuration hypotheses. Yet, computing the optimal action that  maximizes the information gain involves integral over all latent variables, which is intractable. One can approximate this by a sampling-based method~\cite{interactiveperception}. However, the high computational requirements still prohibit the agent from solving the task within a reasonable time. We address this by using only a single particle to make a noisy approximation of the optimal action.%, which is even more efficient than the sampling-based method  in \cite{interactiveperception}.
%\AC{It may be useful to formally define what the uncertainty is in the setup, and what is meant by reducing the uncertainty using an action.} 
%We utilize virtual simulations to generate an optimal action that can reduce uncertainty of the joint configuration hypotheses. Yet, computing optimal action that  maximizes optimal information gain involves integral  over all latent variables, which is intractable. In \cite{interactiveperception}, an approximate version using a sampling-based method is proposed, however the high computational requirements may prohibit the agent from solving the task within reasonable time. To address this problem, we heuristically use only a single particle to make a noisy approximation of the optimal action.

We sample a joint configuration from the set of particles $s^{(k)} \sim \mathcal{S}$ and obtain the optimal action by simulating different actions on the object with the physics simulation. The action $a_t = (p, r) \in\mathbb{R}^6$ is represented as a 3D point $p_t \in \mathbb{R}^3$ on the object and the direction $r_t \in \mathbb{R}^3$ to apply force.
The optimal action is defined as the action that can maximally deform the object or a target object part over a single step. For multi-part objects, we maintain a list of parts-of-interest, which we will introduce shortly, and we sample a target part from the list to act on. We measure how much an object part $i$ deforms by $d_i = \| \theta_{t+1}^{\text{cur}_i} - \theta_{t}^{\text{cur}_i} \|.$ Although one can naively sample a huge number of actions and pick the best action through simulation, we found this can be extremely inefficient with large object parts. To improve inference speed, we instead treat the action inference as a particle filtering problem: we initialize a number of action proposals by randomly sampling 3D locations on the target point cloud and assign random directions to apply force, then we use the measured distance $d_j$ as the likelihood to update the posterior distribution of the particles. 
We add noise to the action while reproducing the particles from previous iterations.
We continue this process \icra{three} times and finally sample a particle from the pool to obtain the action $a^\ast$. 
\footnote{\icra{We found the particle filter (PF) generates nearly optimal action $1,500$ times faster compared to an oracle optimal action generated by exhaustive search (ES). We compare deformations caused by them and found PF with $100$ particles almost always generates the same action ($d_i^\text{PF}/d_i^\text{ES}=0.995$).}}
We found the inferred action $a^\ast$ is often close to the oracle optimal action that maximizes $d_i$.
% We evaluate the efficiency and optimality of this action selection method in \sref{subsec:exp_pf_act}.
%the most probable particle as an optimal action and position $a^\ast,p^\ast = s^\text{act}_i$, where $i = \argmax_{i\in{1,...,N^\text{act}}} d_i$ to be applied to the test object at the next step. 

The probabilistic formulation of an articulation mechanism given past observation and action is
\begin{equation}
    \begin{split}
        p(s_t | O_{1:t-1}
        &, a_{1:t-1}) \\ 
        &= \int \underbrace{p(s_t | s_{t-1}, a_{t-1})}_{\text{forward dynamics}} \underbrace{p(s_{t-1} |  O_{1:t-1}, a_{1:t-1})}_{\text{obtain through recursion}},
    \end{split}
    \label{eq:pred}
\end{equation}
where the first term is handled by the physics engine by forward simulation, and the second is initialized with the prior defined in Eq.~\eqref{eq:prior} and can be obtained through recursion.

%\AC{It would look better if you have some math for the posterior distribution.}
%\konote{Will entirely update this section to include more maths to describe the generative model.}

\subsection{Updating hypotheses through analysis-by-synthesis} \label{subsec:update_hypotheses}
%\AC{How is the domain gap between simulation and real world handled? } \konote{Will write some potential difficulties and how to resolve that in the limitation section later.}
We apply the inferred action $a^\ast$ on the target object $O_t$  
%in the real world
to observe outcome $O_{t+1}$. We then update the probability of each hypothesis through analysis-by-synthesis: we first apply the same action $a^\ast$ on all the "imagined" objects, $s \in \mathcal{S}$ in the physics engine. After applying the action, we obtain $\hat{O}^{(k)}_{t+1}$ for each particle $s^{(k)}.$
%we first load the joint configuration, and then apply the action -- and observe the pointcloud $o_{t+1}$.
We define the likelihood of the particle $s^{(k)}$ as $w_k = \frac{1}{\text{dist} (O_{t+1}, \hat{O}^{(k)}_{t+1}) + \epsilon}$, where $\text{dist}(o_1, o_2) = \frac{1}{|o_1|} \sum_{x \in o_1} \min_{y \in o_2} \| x - y \|_2^2$ is the chamfer distance between two point cloud $o_1$ and $o_2.$ 
%\AC{Shouldn't $w_k$ be some form of softmax? if dist() is close to zero, you will have close to $\infty$ for  $w_k$; however this needs to be normalized across different particles to form a distribution. there must be something like $w_k = w_k/\sum_i w_i$?.} 
The overall updated posterior is:
% \begin{equation}
% p(s_t | O_{1:t}, a_{1:t-1}) \propto  p(O_t | s_{t}) p(s_t|O_{1:t-1}, a_{1:t-1}) = \sum_{k=1}^K w_k p'(s_t|O_{1:t-1}, a_{1:t-1}), \label{eq:posterior}
% \end{equation}
\begin{equation}
	\begin{split}
	    p(s_t | O_{1:t}, a_{1:t-1})
	    &  \propto p(O_t | s_{t}) p(s_t|O_{1:t-1}, a_{1:t-1})\\
	    & = \sum_{k=1}^K w_k p(s_t|O_{1:t-1}, a_{1:t-1}),
    	\label{eq:posterior}
	\end{split}
\end{equation}
where the second term can be computed from Eq.\eqref{eq:pred}, and the whole inference is implemented through particle filtering with weighted sampling. 

% So, we keep updating the particles until we will be confident enough, like more than 90% of particles belong to one specific kinematic tree, and returns the answer.

% \newcommand{\xdifftrain}{\bar{\bm{x}}_t^{\text{train}, i}}
% \newcommand{\xdifftest}{\bar{\bm{x}}^\text{test}_t}

% \begin{equation}
%     w_t^{\text{CD}, i} = 
% \end{equation}

% \begin{equation}
%     w_t^{\text{cos}, i} = 
%     \begin{cases}
%         0 & | \xdifftest | < \epsilon \,\land\, |\xdifftrain| > 0 \\
%         0 & | \xdifftest | > 0        \,\land\, |\xdifftrain| < \epsilon \\
%         \dfrac{ \frac{\xdifftest \cdot \xdifftrain}{|\xdifftest| | \xdifftrain |} + 1 }{2} & \text{otherwise}
%     \end{cases},
%     \text{where} \, \bar{\bm{x}}_t = \bm{x}_{t+1} - \bm{x}_t
% \end{equation}

\begin{figure*}
    \centering
    \includegraphics[width=0.8\textwidth]{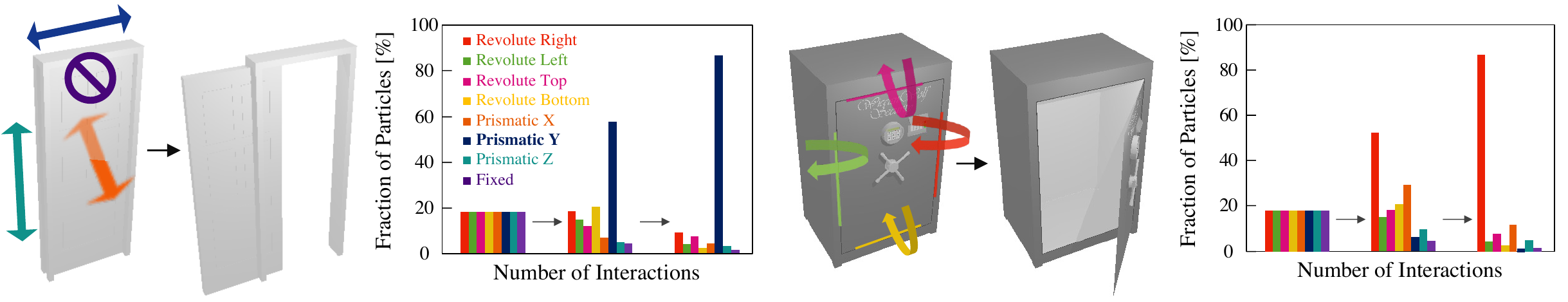}
    \captionof{figure}{Percentage of particles representing different hypothetical configurations during interactions. The hypotheses start as a uniform distribution, then with an observed action, belief tends to aggregate on the correct joint type.}
    \label{fig:joint_type_est}
    \vspace{-2mm}
\end{figure*}

\begin{table*}[t]
	\begin{minipage}{0.69\linewidth}
    \centering
    \resizebox{\textwidth}{!}{
    \begin{tabular}{cccccccccccccc}
        \toprule
                   & & \multicolumn{6}{c}{Novel instances in training Categories}                                           & \multicolumn{4}{c}{Testing categories} \\
                   \cmidrule(l{3mm}r{1mm}){3-8} \cmidrule(l{3mm}r{1mm}){9-12}
                   & & Box          & Door        & Microwave    & Fridge      & Cabinet     & Mean      & Safe        & Table       & Washing     & Mean\\ \midrule
        \multirow{3}{*}{\textit{Closed}}
        & PN2 & $\mb{100.0}$ & $43.3$      & $97.4$      & $72.9$      & $69.2$     & $76.5$      & $55.7$      & $56.5$      & $45.2$       & $54.0 $ \\
        & Ours       & $\mb{100.0}$ & $\mb{85.4}$ & $\mb{100.0}$  & $\mb{98.6}$ & $96.7$& $\mb{96.1}$ & $89.7$ & $98.7$ & $\mb{100.0}$ & $96.1$ \\
        & PN2+Ours& $\mb{100.0}$ & $80.5$ & $90.9$ & $\mb{98.6}$ & $\mb{97.7}$      & $93.4$ & $\mb{96.6}$ & $\mb{99.3}$ & $93.8$ & $\mb{96.5}$\\
        \midrule
        \multirow{3}{*}{\textit{Half-opened}}
        & PN2 & $\mb{100.0}$ & $87.5$      & $\mb{100.0}$& $99.1$       & $\mb{99.8}$&$\mb{97.4}$ & $\mb{100.0}$ & $92.2$      & $77.4$       & $89.9$\\
        & Ours       & $\mb{100.0}$ & $\mb{97.6}$ & $81.8$      & $\mb{100.0}$ & $99.0$     &$95.7$      & $\mb{100.0}$ & $92.8$ & $\mb{100.0}$ & $\mb{97.6}$\\
        & PN2+Ours& $92.3$ & $90.2$ & $\mb{100.0}$ & $98.6$ & $98.6$ & $96.0$ & $\mb{100.0}$ & $\mb{93.4}$ & $93.8$ & $95.7$\\
        \bottomrule
    \end{tabular}}
    \vspace{1mm}
    \caption{Joint type estimation accuracy {[}\%{]}.}
    \label{tab:joint_type_est}
    \end{minipage}
    \begin{minipage}{0.01\linewidth}
    \end{minipage}
	\begin{minipage}{0.30\linewidth}
		\centering
		\includegraphics[width=0.95\textwidth]{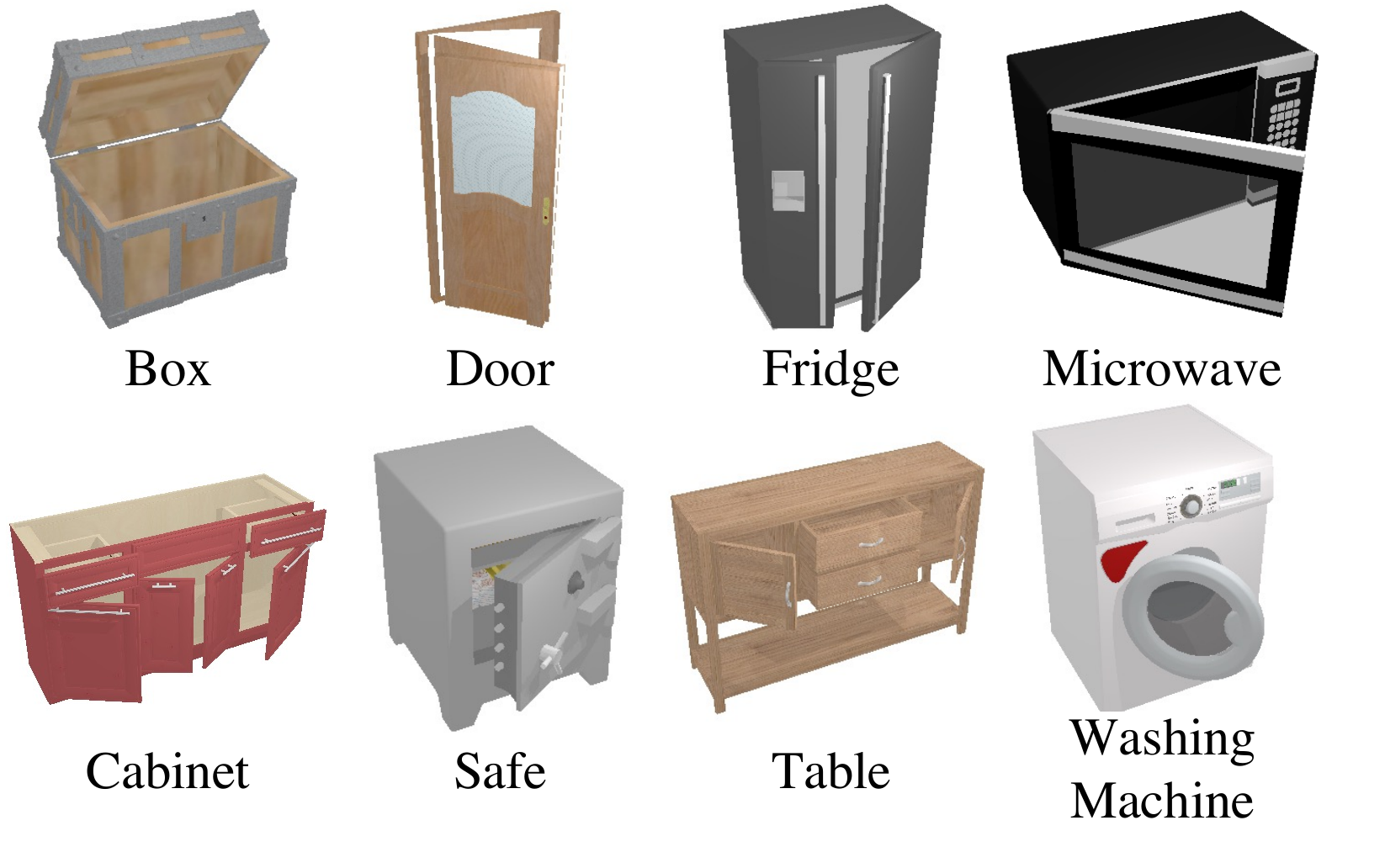}
        \vspace{-3mm}
		\captionof{figure}{{\small Categories from PartNet-Mobility dataset used for our experiments.} }
		\label{fig:categories}
	\end{minipage}
    \vspace{-3mm}
\end{table*}

\subsection{Handling Joints with Dependency in Goal-Conditioned Manipulation} \label{sec:goalmanip}
A real puzzle box often consists of joints with dependencies, e.g., a lock needs to be open first in order to operate on another lock.
%The algorithm we have proposed so far resolves the problem by randomly select a part and act on it, which 
Randomly selecting a part to act on is ineffective and may not be sufficient to solve the problem since (1) the agent can act on a segment that is irrelevant to the task, e.g., a decoration on the box, and (2) the agent can underestimate the joint limit by ignoring the possibility that another part is blocking the current joint. To resolve this issue, we propose to keep track of the relevant parts and gradually grow a dependency tree throughout the exploration process. 
%dependency when a part can not be moved to the desired position due to collision.
%we try to maintain a parts-of-interest list $q_{POI}$ through out the exploration steps, and keep track of the dependency when a part can not be moved to the desired position due to collision.

Given goals in the form of ``moving part X towards Y'', we maintain a parts-of-interest list $q_\text{POI}$ to keep track of task-relevant object parts and their desired position. For example, \icra{consider a door with a few locks, whose}
% in the \textit{PuzzleBoxes} dataset, which consists of a door and a few locks (\fref{fig:puzzle_box}), 
goal is to pull open the door. Thus, we initialize $q_\text{POI}$ by adding the ``door'' part, $O_0$, and the desirable moving direction $d_0$. When selecting an action (see section \ref{sec:action}), 
we always act on the most recently added object part. In the first run, we select the door since it is the only part in the list.
% we always select an action to act on the object part that most recently get added to the list, e.g., in the first run, we always select the door since that is the only part in the list. 

Using the physics engine, we not only infer the optimal action that would cause desirable changes to the target part, but also detect object parts, e.g., \icra{locks on the \textit{PuzzleBoxes} we introduce shortly}, that will collide with it. We consider these collided parts as having a dependency with target part at hand. We can further infer the desirable change direction $d_i$ for each of these collided parts $O_i$ that would unblock the current part. Then, we add the part along with the desired changing direction to $q_\text{POI}.$ Sometimes multiple directions might lead to a successful unblock, in this case, we randomly select one direction to be put in the list. We expect the pool of particles to keep track of different sampling outcomes. We can keep adding ``unsolved'' parts with dependencies to the current parts to the list. A part is marked as ``solved'' and removed from the list if it can be and has been changed to a desired configuration that unblocks its parent node in the dependency tree. 
% Please refer to the supplementary material for the complete algorithm, and \sref{subsec:puzzleboxes} for more details about PuzzleBoxes.
% Please refer to \sref{subsec:puzzleboxes} for more details about PuzzleBoxes.
    \begin{table*}[h!]
	\begin{minipage}{0.66\textwidth}
	\centering
	\resizebox{\textwidth}{!}{%
    \begin{tabular}{lcccccccccc} \toprule
        & \multicolumn{6}{c}{Novel instances in training categories} & \multicolumn{4}{c}{Testing categories} \\
        \cmidrule(l{3mm}r{1mm}){2-7} \cmidrule(l{3mm}r{1mm}){8-11}
        Method          & Box    & Door & Microwave & Fridge & Cabinet & Mean &  Safe & Table & Washing & Mean \\ \midrule
        \multicolumn{11}{c}{Binary classification accuracy [\%] $\uparrow$} \\ \midrule
        %W2A ($0.1$K)    & $52.2$ & $49.9$ & $55.0$ & $47.6$ & $49.7$ & $50.9$ && $45.7$ & $48.0$ & $44.7$ & $46.1$\\
       % W2A ($1$K)      & $45.7$ & $54.8$ & $71.6$ & $63.2$ & $62.8$ & $59.7$ && $60.9$ & $65.5$ & $54.6$ & $60.4$\\
        W2A ($10$K)     & $68.6$ & $59.7$ & $70.6$ & $70.1$ & $69.7$ & $67.7$ & $68.5$ & $63.9$ & $60.3$ & $64.2$ \\
        W2A ($100$K)    & $75.9$ & $60.2$ & $81.1$ & $71.3$ & $70.0$ & $71.7$ & $74.1$ & $53.5$ & $66.1$ & $64.6$\\
        Ours ($0.01$K) & $\mb{96.4}$ & $\mb{79.5}$ & $\mb{97.8}$ & $\mb{93.0}$ & $\mb{93.0}$ & $\mb{91.9}$ & $\mb{93.3}$ & $\mb{97.4}$ & $\mb{91.3}$ & $\mb{94.0}$\\ \midrule
        \multicolumn{11}{c}{Distance prediction error $\downarrow$} \\ \midrule
        %W2A ($0.1$K)    & $0.103$ & $0.093$ & $0.068$ & $0.082$ & $0.076$ & $0.084$ && $0.071$ & $0.053$ & $0.097$ & $0.074$ \\
       % W2A ($1$K)      & $0.089$ & $0.084$ & $0.045$ & $0.076$ & $0.076$ & $0.074$ && $0.062$ & $0.053$ & $0.083$ & $0.066$\\
        W2A ($10$K)     & $0.051$ & $0.074$ & $0.040$ & $0.068$ & $0.062$ & $0.059$ & $0.057$ & $0.053$ & $0.076$ & $0.062$\\
        W2A ($100$K)    & $0.049$ & $0.072$ & $0.032$ & $0.063$ & $0.057$ & $0.055$ & $\mb{0.051}$ & $0.061$ & $0.067$ & $0.059$\\
        Ours ($0.01$K) & $\mb{0.009}$ & $\mb{0.036}$ & $\mb{0.013}$ & $\mb{0.040}$ & $\mb{0.026}$ & $\mb{0.025}$  & $0.055$ & $\mb{0.016}$ & $\mb{0.029}$ &$\mb{0.033}$ \\ \bottomrule
    \end{tabular}}
    \vspace{1mm}
    \caption{Affordance prediction performance.}
    \label{tab:affordance}
	\end{minipage}\hfill
	\begin{minipage}{0.02\textwidth}
	\end{minipage}
	\begin{minipage}{0.32\textwidth}
		\centering
		\includegraphics[width=0.95\textwidth]{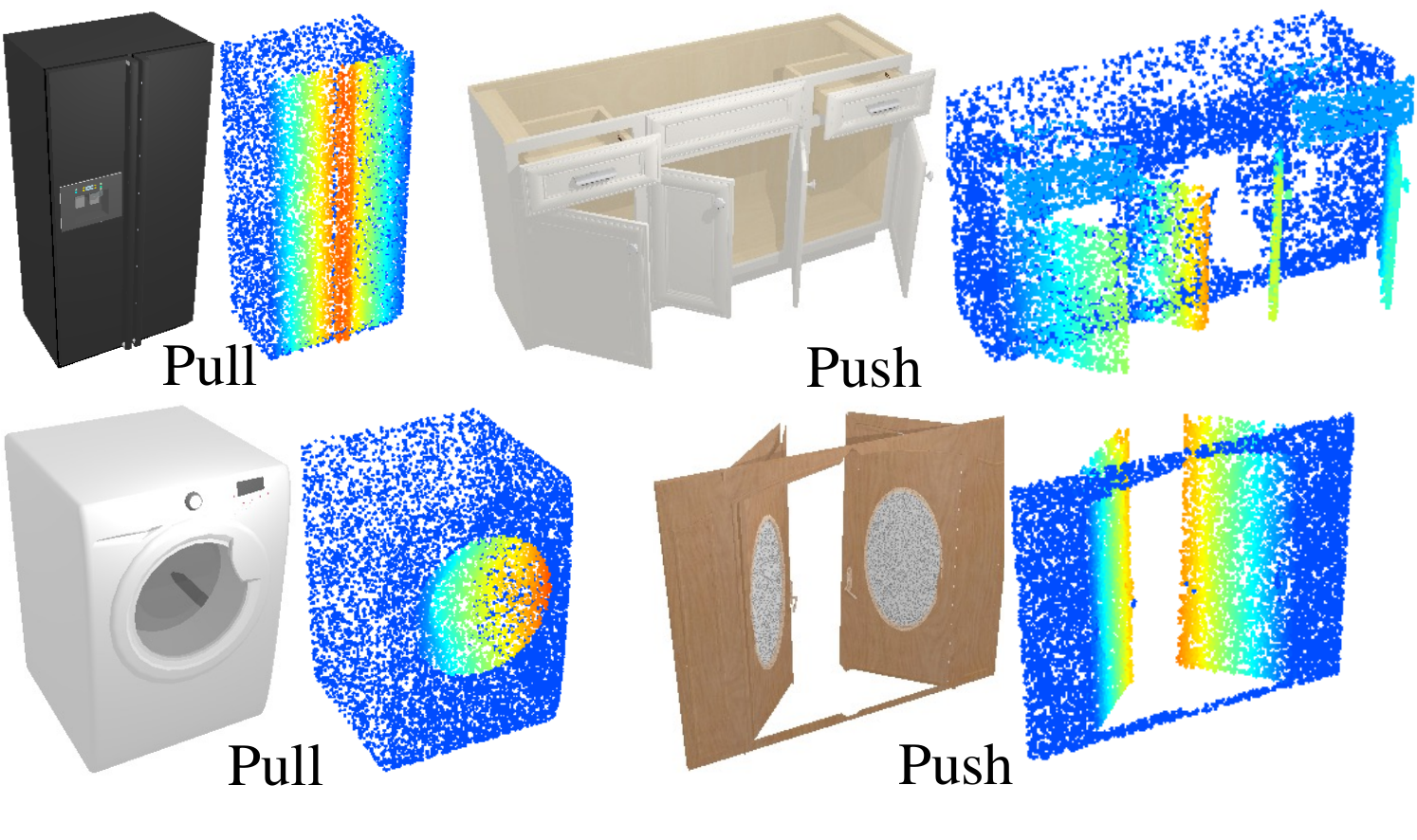}
        \captionof{figure}{{\small
        % Qualitative visualizations of distance prediction. 
        \icra{Visualizations of distance prediction. 
        % The left two are affordance maps of pulling forces, and the right two pushing forces. 
        The warmer color shows larger deformations.}}}
    \label{fig:affordance}
	\end{minipage}
    \vspace{-2mm}
\end{table*}

\section{Experiments}
% In the experiments, we want to answer the following four questions.
% \begin{itemize}
%     \item Can interactions improve joint type estimation performance?
%     \item Can our method predict affordance maps more sample efficient than other interaction based approaches?
%     \item Can our method manipulate articulated objects more sample efficient?
%     \item Can our method interact with articulated objects that have joint chains?
% \end{itemize}
% First, we want to demonstrate that we can improve joint type classification performance by interacting with the object.
% Then we want to know whether our approach can generate an accurate affordance map, which tells us where to interact.
% And then we want to know whether our approach improves sample efficiency when manipulating the articulated objects.
% And finally we will show the performance on more complex settings, where articulated objects have joint chains.
\icra{
We evaluate \modelname~on both the PartNet-Mobility dataset and PuzzleBoxes dataset on SAPIEN~\cite{xiang2020sapien} physics engine.}

The \textbf{PartNet-Mobility} \icra{dataset provides a wide variety of synthetic articulated objects. We specifically use $8$ different categories as shown in~\fref{fig:categories} with two different settings: In the \textit{closed} setting, all movable joints are shut, which is often the most visually ambiguous setup for an object. In the \textit{half-opened} setting, all joints are initialized at the midway point between the joint limits.}

The \textbf{PuzzleBoxes} \icra{dataset has more challenging configurations and joint dependency.
Inspired by the puzzle box experiment by Thorndike \cite{Thorndike1911-THOAI}, we manually design PuzzleBoxes with different levels with different number of \textit{locks} ($N^\text{locks}$) and dependency \textit{chains} ($N^\text{chain}$).
As shown in \fref{fig:puzzle_box}, we prepared five different settings: $(N^\text{chain}, N^\text{locks}) \sim \{(1,1),(2,1),(3,1),(1,2),(1,3) \}$, where each setting has 10 different configurations (joint type, axis, and position).}
% The dataset consists of locked boxes that requires unlocking multiple joints with dependency as shown in \fref{fig:puzzle_box}.

In both dataset, the 6-DoF action is implemented in the simulator by simulating a directed force on a 3D point, imitating actions from a suction gripper.

\subsection{Joint type estimation} \label{sec:exp_jnt_est}
%We first evaluate our method's effectiveness in estimating joint type. 
We first evaluate how well \modelname can estimate the type, location, and limits of joints on an articulated object. 
%The task is to find joint types of all movable joints given an articulated object.

\textbf{Settings.}
% We test joint estimation using the PartNet-Mobility dataset, with two different settings for initializing the object. In the \textit{closed} setting, all movable joints are shut, 
% which is often the most visually ambiguous setup for an object. In the \textit{half-opened} setting, all joints are initialized at the midway point between the joint limits.
We test joint estimation using the PartNet-Mobility dataset with both the \textit{closed} and \textit{half-opened} settings.
%: $\theta_{t=0} = (\theta^\text{high} - \theta^\text{low})/2$. 
To measure the performance, we cast the problem into an eight-way classification problem where the model classifies the target joint as one of the followings: four different revolute joints attached to the right, left, top, or bottom of the 3D bounding boxes for the object part, three different prismatic joints that moves along each of the X, Y, and Z axes, or a fixed joint (see \fref{fig:joint_type_est}). 

%different possible hypotheses for the joint configuration, which spanned the range of joint types included in the dataset:
%For the joint configuration hypotheses, we empirically found that eight different joint types can cover the whole categories -- 
%four revolute joints attached to the right, left, top, or bottom of the 3D segments of the object part, three prismatic joints that moves the target link along each of the X, Y, and Z axes, and one fixed joint. 

\textbf{Models.}
We initialize a uniform prior for \modelname~ with the eight possible joints, using $110$ particles.
The algorithm stops if one of the following conditions is satisfied: (1) the model has good confidence with more than $90\%$ of the particles belong to a single class, or (2) the model interacts with the object $10$ times.
%The \modelname~model assumed eight different possible hypotheses for the joint configuration, which spanned the range of joint types included in the dataset:
%For the joint configuration hypotheses, we empirically found that eight different joint types can cover the whole categories -- 
%four revolute joints attached to the right, left, top, or bottom of the 3D segments of the object part, three prismatic joints that moves the target link along each of the X, Y, and Z axes, and one fixed joint. We initiate a total of 113 particles for our method.
%For each joint type $c$, we prepare two opposite joint limits to cover the object that moves both directions. Also, we use $8$ particles per each joint type and joint direction, which are evenly spaced numbers over $[0, 1]$. Therefore, the total number of particles will be $2 \times 8 \times 7  + 1 = 113$, where $1$ is for fixed joint.
We compare our algorithm to a supervised learning baseline \textit{PN2}~\cite{articulated_li}, which uses PointNet++~\cite{qi2017pointnet++} as a feature extraction backbone
to predict joint types given an input point cloud and segmentation masks of the object parts connected to the joint. We train the model to classify the link as one of the eight joint types described above.
%, and does not explore in test time. The PN2 takes the concatenation of the pointcloud of the target object and segmentation mask that specifies a target movable link which we want to estimate its joint type, and classifies the link to eight different joint types described above. To train the model, we collect $100$K data by randomly initializing the joints states of all movable joints.
We also test the combination of \modelname~and PointNet++, which we denote \textit{Ours + PN2}, where we use the trained PointNet++ model as a prior when initializing particles. 
% We describe more details in supplementary materials.

\textbf{Results and analysis.}
\Tref{tab:joint_type_est} shows the joint type estimation accuracy.
Our model performs comparably with PN2 in the \textit{half-opened} setting and significantly outperform PN2 on the \textit{closed} setting where joint type is mostly ambiguous from vision alone. We also show, by integrating visual prior from PN2 with the proposed framework, we can improve in cases where visual prior helps  significantly, e.g., in \textit{half-opened} Microwave.
%Looking at the big performance gap between ours and PN2 on the \textit{closed} setting, we can see the interaction resolves the visually ambiguous situations. We can also see the effectiveness of interactions by comparing the results of PN2 and PN2 + Ours, the low performance of PN2 are improved by testing-time interaction.
% We also see our method consistently performs high accuracy even in the testing categories, because our method requires zero-training. 
%For \textit{Half-Opened} settings, the baseline performs comparably to ours, since it is visually easier to identify the joint type.
We visualize the posterior over hypotheses in \fref{fig:joint_type_est}. We can see our model becomes more confident after a few interactions.
% To further evaluate the joint type estimation performance under stochastic dynamics, we show our method is robust to action noises in~\sref{subsec:exp_stochastic_dynamics}. Also, we compare the performance on different scoring functions for updating hypotheses in \sref{subsec:chamfer_vs_cos}.

\subsection{Joint type estimation under stochastic dynamics.} \label{subsec:exp_stochastic_dynamics}
\textbf{Settings.}
We further evaluate the performance of H-SAUR under stochastic dynamics by adding noise to action to imitate a stochastic dynamics. The action noise $\epsilon$ is uniformly sampled from $\bm{\epsilon} \sim [-\sigma, \sigma]^6$ thus the action on the stochastic dynamics will be $a^\text{noise}_t = a_t + \bm{\epsilon}$. We evaluate H-SAUR with $\sigma \in \{ 0, 0.1, 0.2, 0.3 \}$, where $\sigma=0$ corresponds to \textit{Ours} in \Tref{tab:joint_type_est}. 

\textbf{Results and analysis.}
\Tref{tab:joint_type_est_ablation_noise} shows the results on different noise levels.
At its most extreme, the noise is sampled from a uniform range of width $0.6$ meters, which is the equivalent to the size of the articulated part in many cases, yet adding this noise has little effect on joint estimation performance. This is partly because our method is a probabilistic framework, thus it can handle any uncertainty including stochastic dynamics, part segmentation, action noises, etc.

\begin{table}[t]
    \small
    \centering
    \vspace{-2mm}
    \resizebox{\columnwidth}{!}{
    \begin{tabular}{cccccccccccccc}
        \toprule
        $\sigma$           & Box          & Door        & Microwave    & Fridge      & Cabinet     & Safe        & Table       & Washing     & Mean\\ \midrule
        $0.0$   & $\mb{100.0}$ & $85.4$ & $\mb{100.0}$ & $\mb{98.6}$ & $96.7$& $89.7$ & $\mb{98.7}$ & $\mb{100.0}$ & $\mb{96.1}$ \\
        $0.1$  & $96.1$ & $\mb{94.8}$ & $\mb{100.0}$ & $87.8$ & $90.9$ & $\mb{95.9}$ & $96.6$ & $87.5$ & $93.6$ \\
        $0.2$  & $95.4$ & $92.8$ & $\mb{100.0}$ & $82.9$ & $\mb{100.0}$ & $\mb{95.9}$ & $82.8$ & $87.5$ & $92.2$ \\
        $0.3$  & $95.2$ & $92.8$ & $\mb{100.0}$ & $85.4$ & $\mb{100.0}$ & $93.2$ & $96.6$ & $93.8$ & $94.6$ \\
        \bottomrule
    \end{tabular}
    }
    \caption{{\small Joint type estimation accuracy on noisy dynamics {[}\%{]}.}}
    \label{tab:joint_type_est_ablation_noise}
    \vspace{-4mm}
\end{table}

\begin{table*}[t]
	\begin{minipage}{0.62\textwidth}
    \centering
    \resizebox{\textwidth}{!}{
    \begin{tabular}{lcccccccccc}
        \toprule
        & \multicolumn{6}{c}{Novel instances in training categories} & \multicolumn{4}{c}{Testing categories} \\
        \cmidrule(l{3mm}r{1mm}){2-7} \cmidrule(l{3mm}r{1mm}){8-11}
                          & Box    & Door    & Microwave & Fridge & Cabinet & Mean   & Safe    & Table & Washing & Mean \\ \midrule
        %W2A ($0.1$K)      & $1.7$  &  $25.6$ & $0.0$     & $0.2$  & $0.9$   & $5.7$  && $0.1$ &  $0.1$ &  $0.1$ & 0.1 \\
        %W2A ($1$K)        & $0.1$  &  $30.1$ & $87.2$    & $40.7$ & $52.1$  & $42.0$ && $31.1$ &  $82.2$ &  $63.4$ & 58.9\\
       % W2A ($10$K)       & $36.5$ &  $58.1$ & $51.3$    & $55.2$ & $76.6$  & $55.5$ && $43.3$ &  $71.4$ &  $71.0$ & 61.9\\
        W2A ($100$K)      & $65.0$ &  $47.3$ & $49.6$    & $50.7$ & $54.8$  & $53.5$ & $42.6$ &  $43.9$ &  $70.0$ & 52.2\\
        %W2A + HP ($0.1$K) & $1.3$  &  $25.4$ & $0.1$     & $0.2$  & $1.0$   & $5.6$  && $0.5$ &  $7.5$ &  $1.1$ & 3.0\\
        %W2A + HP ($1$K)   & $0.1$  &  $32.0$ & $99.8$ &  $36.3$ &$60.3$ &$45.7$ &&  $29.9$ &  $89.4$ &  $67.4$ & 62.2\\
        %W2A + HP ($10$K)  & $76.4$ &  $49.4$ & $81.8$ &  $90.6$ &  $86.4$ & $76.9$   &&  $67.7$ &  $70.5$ &  $\mb{94.6}$ & 77.6\\
        W2A+HP ($100$K) & $\mb{96.2}$  & $51.3$ &  $91.0$ &  $\mb{98.4}$ &  $92.9$ &$85.9$ &  $72.8$ &  $72.0$ &  $94.3$ & 79.7\\
        Ours ($0.01$K)   & $83.9$ &  $\mb{90.2}$ & $99.0$ &  $94.8$ &  $95.6$ & $92.7$ &  $82.8$ &  $98.0$ &  $93.8$ & $91.5$ \\
        Ours + PN2        & $87.2$ &  $86.0$ &  $\mb{100.0}$  &  $94.7$ &  $\mb{97.1}$ & $\mb{93.0}$ & $\mb{85.6}$ &  $\mb{99.0}$ &  $\mb{97.7}$ & $\mb{94.1}$ \\
        % Ours + PN2 (M)    &  $64.8$ &  $80.3$ &  $89.5$ &  $80.5$ &  $89.1$ & $80.8$ && $47.7$ &  $97.5$ &  $93.7$ & $79.6$ \\
        % Ours + PN2 (S)    & $43.6$ &  $60.9$ &  $30.8$ &  $44.2$ &  $62.5$ & $48.4$ &&  $24.7$ &  $78.8$ &  $61.4$ & $55.0$ \\
        % ANSCH  &  &  &  &  &   &  &  &  \\ 
        \bottomrule
    \end{tabular}}
    % \vspace{-2mm}
    \caption{The proportion of the part opened [\%] with fifteen testing-time interactions.}
    \label{tab:manipulation}
	\end{minipage}\hfill
	\begin{minipage}{0.02\textwidth}
	\end{minipage}
	\begin{minipage}{0.36\textwidth}
        \centering
	    \vspace{1mm}
        \resizebox{\columnwidth}{!}{
        \begin{tabular}{lccccccccccc}
            \toprule
            % & \includegraphics[width=1in]{fig/all_puzzle_boxes.png} \\
            \multirow{2}{*}{Setting}   & \multirow{2}{*}{1-chain}    & \multirow{2}{*}{2-chain}    & \multirow{2}{*}{3-chain} & 1-chain & 1-chain \\ 
             & & & & 2-locks & 3-locks \\ \midrule
            Random    & $23.3$ & $6.7$ & $0.0$ & $13.3$ & $3.3$ \\
            Heuristic & $36.7$ & $13.3$ & $0.0$ & $23.3$ & $10.0$ \\
            CURL & $33.3$ & $0.0$ & $0.0$ & $0.0$ & $0.0$ \\
            Ours      & $\mb{96.7}$ & $\mb{86.7}$ & $\mb{80.0}$ & $\mb{93.3}$ & $\mb{86.7}$  \\
            \bottomrule
        \end{tabular}
        }
        \vspace{1mm}
        \caption{Manipulation performance ($\%$) for solving PuzzleBoxes averaged from 3 runs.}
        \label{tab:puzzle_box}
	\end{minipage}
\end{table*}

\subsection{Action Proposal and Affordance Map} \label{sec:affordance}
We next measure how well the 
\modelname~model can use its estimates of joint properties to estimate whether an action will be effective on the PartNet-Mobility dataset.
%the action proposal performance of the proposed method. 
%Given a point in the pointcloud of an articulated object and a force direction, we evaluate (1) the accuracy of forecasting whether the point is effective or not and (2)the l2 distance between the predicted pose and the ground truth.

%future prediction performance.

\textbf{Settings.}
To evaluate all models, we collect $10,000$ interactions on the \textit{closed} setting by randomly sampling a point belonging on a movable part and applying a force with a uniformly distributed direction on the surface of the 3-d unit sphere. An action is labeled as "success" if it causes the joint to move more than 5\% of its full range. We counterbalance "success" and "failure" interactions in the final test set.
%To evaluate the model, we collect $10,000$ pairs of positive and negative interactions on the \textit{closed} setting by uniformly sampling a point and action direction.
We use two metrics to evaluate the models: (1) \textit{Binary classification Accuracy} which is the proportion of actions correctly predicted as success or failure, and (2) \textit{Distance Prediction} which measure the $\ell_1$ distance between the predicted point translation and the ground truth.

\textbf{Baseline.}
We compare our model with the state-of-the-art articulated object manipulation algorithm, Where2Act (W2A)~\cite{mo2021where}, which takes the pointcloud of an articulated object as input to predict an effectiveness score for all points.
%all links as input and selects single-step action. 
To train the model, we collect $\{10\text{K}, 100\text{K} \}$ number of counter-balanced interactions using the same procedure as above. For a fair comparison, we collect both the testing and training data from only movable links by applying a segmentation mask when sampling the position to interact as our method assumes segmentation of the parts is given.

%to train the model by randomly sample initial position, force direction, and open angles for all movable parts. 

\textbf{Results and analysis.}
We show the results in \Tref{tab:affordance}. Our method significantly outperforms the baseline, despite being $1000$ times more sample efficient. We show qualitative results of distance prediction by H-SAUR in \fref{fig:affordance}.

\subsection{Manipulation} \label{sec:manip}
Next we evaluate the estimated joints for manipulation task on the PartNet-Mobility Dataset.

\textbf{Settings.}
The task is to open the movable parts as much as possible from completely closed setting within $N^\text{max}=15$ interactions. Our method uses first $N^\text{int}=10$ interactions to estimate the joint type, and the rest to manipulate the object while the baseline models use all $N^\text{max}$ interactions to open the movable parts.
For evaluation, we measure the proportion of the part opened $r=\max_{t\in \{1,...,N^\text{max} \}}(\theta_i - \theta^\text{init})/(\theta^\text{max} - \theta^\text{init})$, where $r=1$ means fully opened target part.

\textbf{Baselines.}
We again use Where2Act as the baseline for this experiment. We also add Where2Act + HP, which employs an additional heuristic that filters out actions that has a larger than 90 deg angle with last-step action as done in~\cite{xu2022umpnet}. This heuristic helps to avoid sequences of back-and-forth actions.

\textbf{Results and analysis.}
\Tref{tab:manipulation} shows our method significantly outperforms Where2Act in all categories and performs better than Where2Act+HP in most settings except for boxes and fridge. We found these two categories are simpler in the sense that all boxes open in the same direction (upward), and so do the fridges (to the left), so it is easy for Where2Act to overfit to a single action.
We can also see that the performance of our method, when combined with the learned PN2 model, slightly improves. It shows \modelname~ alone is already robust to skewed prior, and one can easily improve its performance by incorporating good prior from vision models.  
%\konote{more explanation}
% \konote{will show accuracy curves (x: number of interactions, y: accuracy.}

\subsection{PuzzleBoxes} \label{subsec:puzzleboxes}
Finally, we evaluate \modelname~on a novel benchmark \textit{PuzzleBoxes}.
% which consists of locked boxes that requires unlocking multiple joints with dependency as shown in \fref{fig:puzzle_box}.
The task is to open a door outward more than $60^\circ$ within $100$ interactions. However, opening this door requires first moving other ``locks'' that restrict the door's range of motion as shown in~\fref{fig:puzzle_box}.
%the objects have joint chains in which some movable links cannot move until some other links move to some extent.
% See \fref{fig:puzzle_box} for some examples.
%shows examples of the proposed PuzzleBoxes dataset.
%For example, the door of the leftmost box cannot be opened until an agent unlocks the yellow lock. Thus, the algorithm needs to equip a sequential decision making ability to solve the proposed dataset.

% \textbf{Dataset Details.}
% Inspired by the puzzle box experiment by Thorndike \cite{Thorndike1911-THOAI}, we manually design PuzzleBoxes with different levels with different number of \textit{locks} $N^\text{locks}$ and different level of dependency \textit{chains} $N^\text{chain}$.
% For example, $N^\text{chain}=1$ has one joint chain, meaning that a movable part can move if another movable part moves. 
% As shown in \fref{fig:puzzle_box}, we prepared five different settings: $(N^\text{chain}, N^\text{locks}) \sim \{(1,1),(2,1),(3,1),(1,2),(1,3) \}$, where each setting has 10 different configurations (joint type, axis, and position). 
%All PuzzleBoxes are hand crafted so that it is not easily solved.
%\konote{put more details}

\textbf{Baselines.}
%learning-based approaches learn action priors strongly depend on the training dataset.
%Since any prior learning-based approaches cannot solve this long-sequence problems
%We assume all models do not have access to the puzzle boxes prior to the test phase, so work that relies on learning action priors from the data.
%Since any prior learning-based approaches cannot solve this long-sequence problems
To the best of our knowledge, none of the prior learning-based approaches can solve this long-term manipulation problem without exhaustively interacting with the objects before deployment time. 
To show this, we train an RL agent with
CURL~\cite{laskin2020curl}, a state-of-the-art image-based RL algorithm. We feed the model with RGBD images as inputs and train the agent with 10K interactions.
% See \sref{subsec:rldetails} for implementation details.
% Thus, instead, we compare 
We also compare our algorithm to two learning-free baselines: (1) \textit{Random:} a policy that uniformly sample a movable link and apply randomly sampled action on it at each time step,
%an action direction, and applies the action to a random point of the specified movable link. 
(2) \textit{Heuristic:} a policy that selects an action in the same way as \textit{Random}, but keeps applying the same action if the object moves at the previous step.
% \konote{Explain why not use w2a.}

% \rb{Add RL baseline.}

\textbf{Results and analysis.}
We show the results in \Tref{tab:puzzle_box}. We can see that the RL baseline trained with 10K timesteps, which corresponds to 100x more timesteps than ours, performs poorly on all but the simplest levels. Deep RL algorithms generally needs enormous amount of interactions to learn, and can fail drastically when the agent is allowed to have a \icra{limited number of interactions}. Aside from poor sample efficiency, most learning-based policies would generate similar actions (drawn from a fixed distribution) given similar observations (as PuzzleBoxes are designed to look similar but have different joint axes), since most policies only take one or a few past observations as inputs and do not update action distributions from past failed interactions.
% Also, since some locks of the PuzzleBoxes are designed to have different joint axes even the positions and orientations are same, an RL agent takes the same observations (RGBD images) but is required to produce different actions. Most RL algorithms cannot deal with this situation because they do not update actions based on the past failure experiences.
%This is mainly due to lack of training data.}
Even baselines that use knowledge of the problem structure (both \textit{Random} and \textit{Heuristic}) perform poorly on all levels. In contrast, \modelname can solve even the most complex levels far above chance. %\rb{while it does not need training data}.

% \begin{figure}[t]
%     \centering
%     \includegraphics[width=\columnwidth]{fig/puzzle_box_examples.png}
%     \caption{Examples from the PuzzleBoxes Benchmark.}
%     \label{fig:puzzle_box}
% \end{figure}
\begin{figure}
    \centering
    \vspace{-2mm}
    \includegraphics[width=\columnwidth]{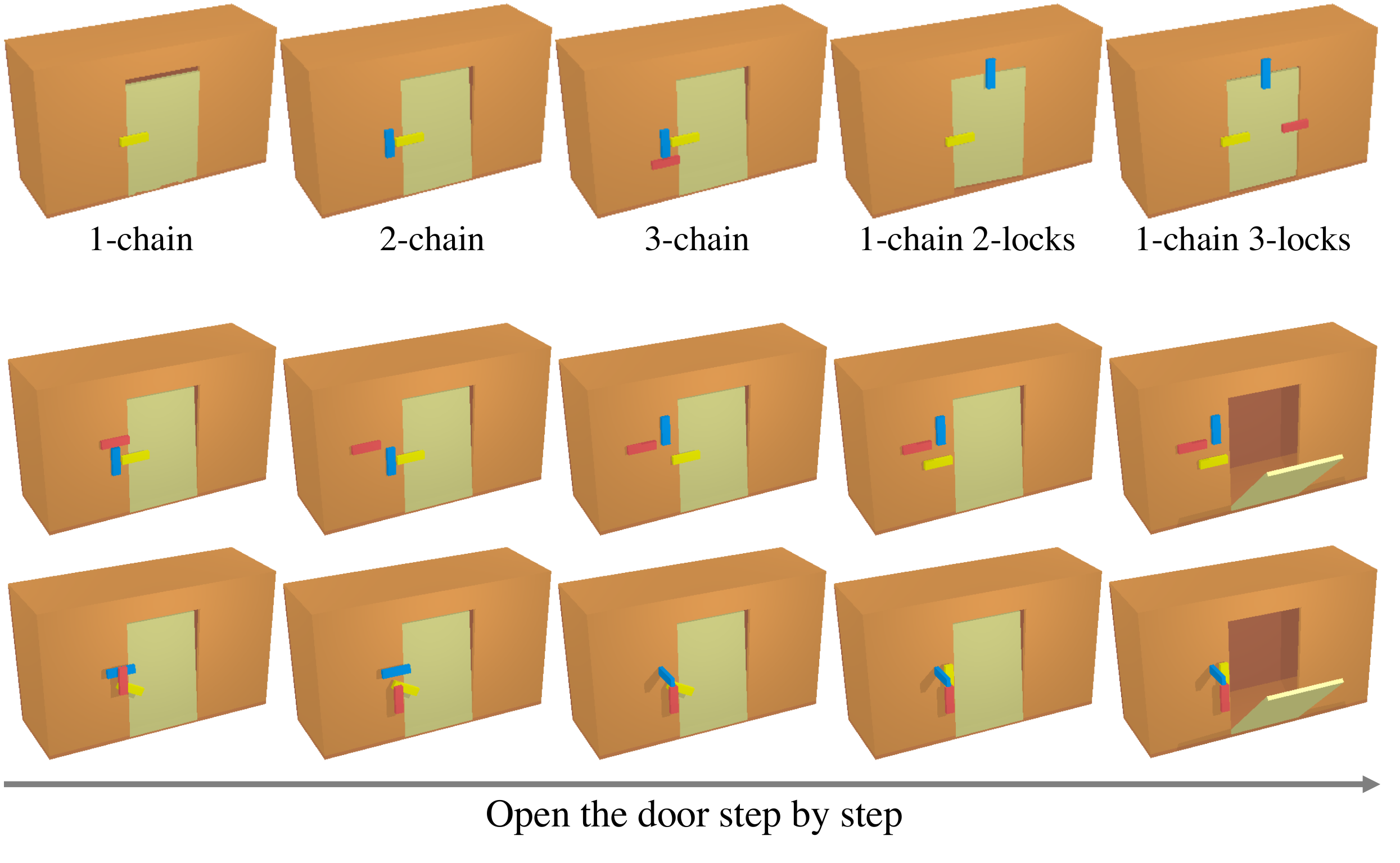}
    \caption{\icra{\textbf{Top} Examples from the PuzzleBoxes Benchmark. \textbf{Bottom} Visualizations of how the puzzle boxes will be opened by unlocking keys in 3-chain puzzle boxes.}}
    \label{fig:puzzle_box}
    \vspace{-2mm}
\end{figure}

    \section{Conclusion}
In this work, we propose a physics- and uncertainty-aware exploration framework, \modelname, that can manipulate diverse articulated objects in circumstances where visual inputs do not uniquely specify the state. We show \modelname~can open complex puzzle boxes requiring several steps to solve. We also show the proposed model outperforms baselines by a large margin, highlighting the importance of quick behavior adaption through test-time exploration. Our model operates directly on pointcloud segments without the needs of detailed tracking using AR tags or any other tracking system, which increases its chance to transfer to a real-world setup. Current results show the model is robust to mismatch between the object of interest and the reconstructed virtual objects from the pointcloud segments.

\icra{
We note that more work 
is needed to extend \modelname to manipulate arbitrary real-world articulated objects.
%To implement our approach on the simulator, we made some assumptions that can potentially limit its application to wider settings.
First, our current model cannot handle articulated joints with arbitrary joint axis, e.g., a door that rotates with a tilted joint, or joint types that have not been prespecified in its hypothesis space. This problem can potentially be addressed using motion-based kinematic prediction \cite{ditto} to propose new hypothesis to include in the prior.
Second, we assume part segments are given. In ongoing work, we are investigating models that jointly infers object parts and their articulations.
%Inferring object parts along with their joint types under the same framework will be an interesting future avenue.
Third, we assume a force can be applied to any point on an object from any direction in order to separate \emph{reasoning} about joints from \emph{manipulating} them. While this is roughly similar to using a suction gripper as in \cite{xu2022umpnet, zeng2020transporter}, we plan to explore practical constraints imposed by a real robot's geometry, gripper, etc. in future work.}

Nonetheless, \modelname demonstrates a promising avenue for systems that can reason about articulated objects, manipulate them, and update beliefs in real time.
%\cite{d3p3} shows 3D scene understanding under similar framework and can potentially be used for part understanding.
%Finally, the current method only handles a limited types of joints. We plan to explore methods that can learn new types of deformation from data.

%which increases it chance to transfer to real world.
%Our model can quickly switch between strategies after failing, by running simulations on ``imagined''hypothetical objects generated by a probabilistic generative model.

%We achieve this with physics-engine-augmented probabilistic generative model where agent can compose hypothetical articulated objects from 3D segments and simulate outcomes given an action. 
    \clearpage
    \bibliographystyle{IEEEtran}
    \bibliography{reference}
    %\clearpage
% \section{Appendix}
\appendix
\subsection{Additional Results}
% \subsubsection{Video}
% To better understand the performance of our framework visually, we include videos of our model as well as the baselines
% solving articulated objects in the PartNet-Mobility Dataset and the PuzzleBoxes Dataset in the {\color{blue} supplementary video}. We also visualize how the hypothesis distribution evolves over time in the {\color{blue} supplementary video}.

\subsubsection{Ablation on different scoring methods to update hypotheses.} \label{subsec:chamfer_vs_cos}
In this section, we compare two different scoring functions for updating hypotheses: chamfer distance as used in our method (described in~\sref{subsec:update_hypotheses}), and cosine similarity referring~\cite{hausman2015active}.
The cosine similarity $\beta_k$ measures the cosine of the angle between the displacement of the real object $\bm{d}$ and the displacement of a hypothetical object. %$\hat{\bm{d}}^{(k)}.$
%$\hat{\bm{d}}^{(k)}$ along which the real object and a hypothetical object moves, respectively.
The direction $\bm{d}$ is computed by $\bm{d}_t = \bm{p}_{t+1} - \bm{p}_t$, where $\bm{p}_t$ is the center position of the observed point cloud $O^j_t$ as $\bm{p}_t = \frac{1}{\| O_t \|}\sum_{\bm{x} \in O^j_t} \bm{x}$.
The cosine similarity is computed by using the direction as $\beta_k = \arccos \left( \frac{\bm{d}_t \cdot \hat{\bm{d}}^{(k)}_t}{|\bm{d}_t||\hat{\bm{d}}^{(k)}_t|} \right)$. 
We then use $\beta_k$ for the likelihood of the $k$-th particle $s^{(k)}$ as $w_k = (\beta_k + 1) / 2$.
This formulation results in an increase of the likelihood when a hypothetical object successfully imitates the movement of the real object.
Following~\cite{hausman2015active}, we assign $w_k=1$ when both objects do not move, and $w_k=0$ in the case of only one of the two objects moves, where we assume such situation can happen when the hypothetical configuration is wrong. To wrap up, cosine similarity-based likelihood function can be formulated as:
\begin{equation}
  w_k =
  \begin{cases}
    (\beta_k + 1) / 2 & \text{if $\bm{d}_t \neq \bm{0}$ and $\hat{\bm{d}}^{(k)}_t \neq \bm{0}$} \\
    1 & \text{if $\bm{d}_t = \bm{0}$ and $\hat{\bm{d}}^{(k)}_t = \bm{0}$}  \\
    0 & \text{otherwise}.
  \end{cases}
\end{equation}

\Tref{tab:joint_type_est_cossim} shows the comparison of the two different scoring functions to update hypotheses for joint type estimation experiments. It clearly shows that chamfer distance outperforms the cosine similarity. We found that cosine similarity works poorly especially when the joint state of the real object is close to the upper or lower limit. In such situations, only either hypothetical or real object moves and results in $w_k = 0$, and filtered out from the particle pool. Chamfer distance, however, is robust to the wrong joint state or joint upper/lower limit, because it still returns a reasonable value even if the estimated state or joint limits is wrong.

\begin{table}[t]
    \small
    \centering
    \resizebox{\columnwidth}{!}{
    \begin{tabular}{ccccccccccccccc}
        \toprule
                   & & Box          & Door        & Microwave    & Fridge      & Cabinet    & Safe        & Table       & Washing     & Mean\\ \midrule
        \multirow{2}{*}{\textit{Closed}}
        & Chamfer distance & $\mb{100.0}$ & $\mb{85.4}$ & $\mb{100.0}$ & $\mb{98.6}$ & $\mb{96.7}$& $\mb{89.7}$ & $\mb{98.7}$ & $\mb{100.0}$ & $\mb{96.1}$ \\
        & Cosine similarity & $78.4$ & $83.0$ & $69.2$ & $73.2$ & $72.7$ & $73.5$ & $74.0$ & $82.8$ & $76.0$ \\
        \bottomrule
    \end{tabular}}
   \vspace{1mm}
    \caption{Comparison of different scoring methods to update hypotheses for joint type estimation accuracy {[}\%{]}.}
    \label{tab:joint_type_est_cossim}
\end{table}

\subsection{Implementation Details}

\subsubsection{Joint Proposals from Part Segments}

\begin{figure}[t]
    \centering
    \includegraphics[width=\columnwidth]{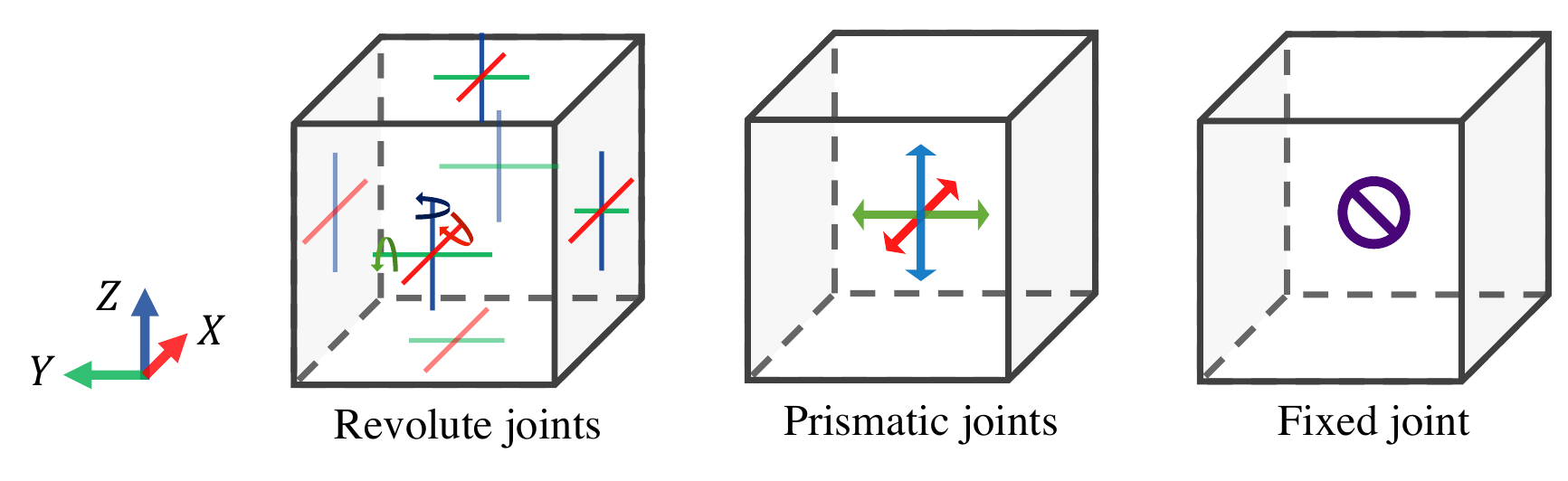}
    \caption{Visualizations of how to define joint positions for the revolute, prismatic, and fixed joints. On the left, you can see X, Y, or Z-axis revolute joints (shown by red, green, and blue colors) are attached to each surface of the axis-aligned bounding box. 
    % On the right, five joint positions for x-axis rotation are shown by cross marks on the projected YZ plane. As for the prismatic joints, we do not change joint positions because we assume the hypothetical object moves towards either x, y, or z-axis direction from its observed position.
    On the middle, we show the prismatic joints that move the hypothetical object towards X, Y, or Z-axis direction from its observed position. Finally, the fixed joint on the right does not change position or orientation of the hypothetical object.
    }
    \label{fig:joints_design}
\end{figure}

Given an object part segment from the point cloud, our method initializes a set of 19 joint proposals from the tight axis-oriented minimum bounding box of the part segment.  
\Fref{fig:joints_design} illustrates the joint proposals from a bounding box.
The 19 joint proposals have their joint positions lie either on the surfaces of the tight bounding box or at the center of the box, and have their joint axes aligned to the X, Y, or Z-axis of the box. Specifically, for revolute joints, we assume the joint lies at the center of one of the 6 surfaces on the box, with 3 possible axes directions aligned with the box, result in $15$ different joint positions ($15 = 3\times 6 - 3$ as there are $3$ duplicate joint positions for each axis). For prismatic joint, we assume the joint lies at the center of the box and can move along 3 axes direction of the box. We found these joint proposals can cover well most articulated objects in the real world.

%We assume joint positions of most movable parts lie either on the surface of the tight bounding box or at the center of the box.
%shows how we attach joint configurations for our hypothetical objects.
%We assume joint positions of most movable parts lie in the surface of the tight bounding boxes.

\subsubsection{Dataset and Simulation Details}

\noindent \textbf{PartNet-Mobility Dataset}
Detailed instance statistics and their joint types for each object category are listed in \Tref{tab:statistics_data_splits}.

\begin{table}[h]
	\centering
	\resizebox{\columnwidth}{!}{%
    \begin{tabular}{ccccccc|ccccc}
        \toprule
        \multicolumn{7}{c}{Training Categories} & \multicolumn{5}{c}{Test Categories} \\ \midrule
        \multirow{2}{*}{Category} & \multicolumn{2}{c}{Instances} & \multicolumn{2}{c}{Joints} & \multirow{2}{*}{Rev.} & \multirow{2}{*}{Pris.} &
        \multirow{2}{*}{Category} & \multirow{2}{*}{Instances} & \multirow{2}{*}{Joints} & \multirow{2}{*}{Rev.} & \multirow{2}{*}{Pris.} \\
                   & Train & Test & Train & Test \\ \midrule
         Box       & 10    & 3    & 10    & 3   & X & - & Safe    & 29  & 29  & X & -\\
         Door      & 26    & 7    & 33    & 8   & X & - & Table   & 62  & 153 & X & X \\
         Microwave & 8     & 3    & 8     & 3   & X & - & Washing & 16  & 16  & X & -\\
         Fridge    & 34    & 9    & 56    & 17  & X & - & \\
         Cabinet   & 269   & 68   & 630   & 157 & X & X & \\ \midrule
         All       & 347   & 90   & 737   & 188 & X & X & All     & 107 & 198 & X & X \\
         \bottomrule
    \end{tabular}}
    \vspace{1mm}
	\caption{Statistics of the data splits.}
	\label{tab:statistics_data_splits}
\end{table}

% We focus on large parts, i.e., ignore small parts such as small buttons to open \textit{Safe}. \todo{this is unclear -- what counts as large and what counts as small? why we exclude them?}
% \konote{Added the following three paragraphs.}
For physics simulation setups, we use frame rate $100$ fps, and all the other settings as default in SAPIEN release. When interacting with an object, we apply the same force for $10$ simulation steps, resulting in $10$ fps simulation. To simulate actions from position control, we set the magnitude of the force by multiplying $100 \times m$, where $m$ is the mass of the target part.
Following Where2Act~\cite{mo2021where}, we disable collision simulation between every pair of two parts connected by an articulated joint. We also set gravity to zero to better simulate articulated objects that have an opener that needs to be opened upwards, e.g., \textit{Boxes}. Setting gravity to zero prevents the opener to gradually fall back from open to close without any force applied.
%, since most objects from ShapeNet~\cite{shapenet2015} are not designed to have accurate geometries that .
%each 
%articulated part to any other object, due to %inaccurate geometry modeling details at joint
%positions, collision shapes for 
%ShapeNet~\cite{shapenet2015} models.
%For physical simulation, we cancel the gravity force to imitate a suction gripper and enable to simulate articulations that open toward Z-axis direction (e.g., \textit{Safe}).

For the rendering settings, we use four cameras to get pointcloud of the target object. We set the near plane to $0.1$, far plane to $100$, resolution to $640 \times 480$, and field of view to $35^{\circ}$. We obtain a pointcloud from the depth images by back-projecting the depth image into pointclouds, removing the far-away (background) points, and down-sampling the point to get a total of $10$K points.
%For getting partial pointcloud, we back-project the depth image into a foreground pointcloud, by rejecting the far-away background depth pixels, and then perform random sampling to get a $10$K-size pointcloud.

For the object parts, we remove parts that are either too small or has function irrelevant to physical articulation. For example, some instances in the \textit{Washing Machine} category have tiny buttons for controlling the machines. We ignore these parts and focus on articulated parts, such as doors.

%focused on visually large enough or meaningful movable parts. For example, instances in \textit{Fridge} category generally have a lot of small buttons to control a fridge (e.g., start or stop washing), but we ignore these small parts and focused on large parts, such as doors.

%apply force to a position ...
% we said it already in the main text

\begin{figure}
    \centering
    \includegraphics[width=\columnwidth]{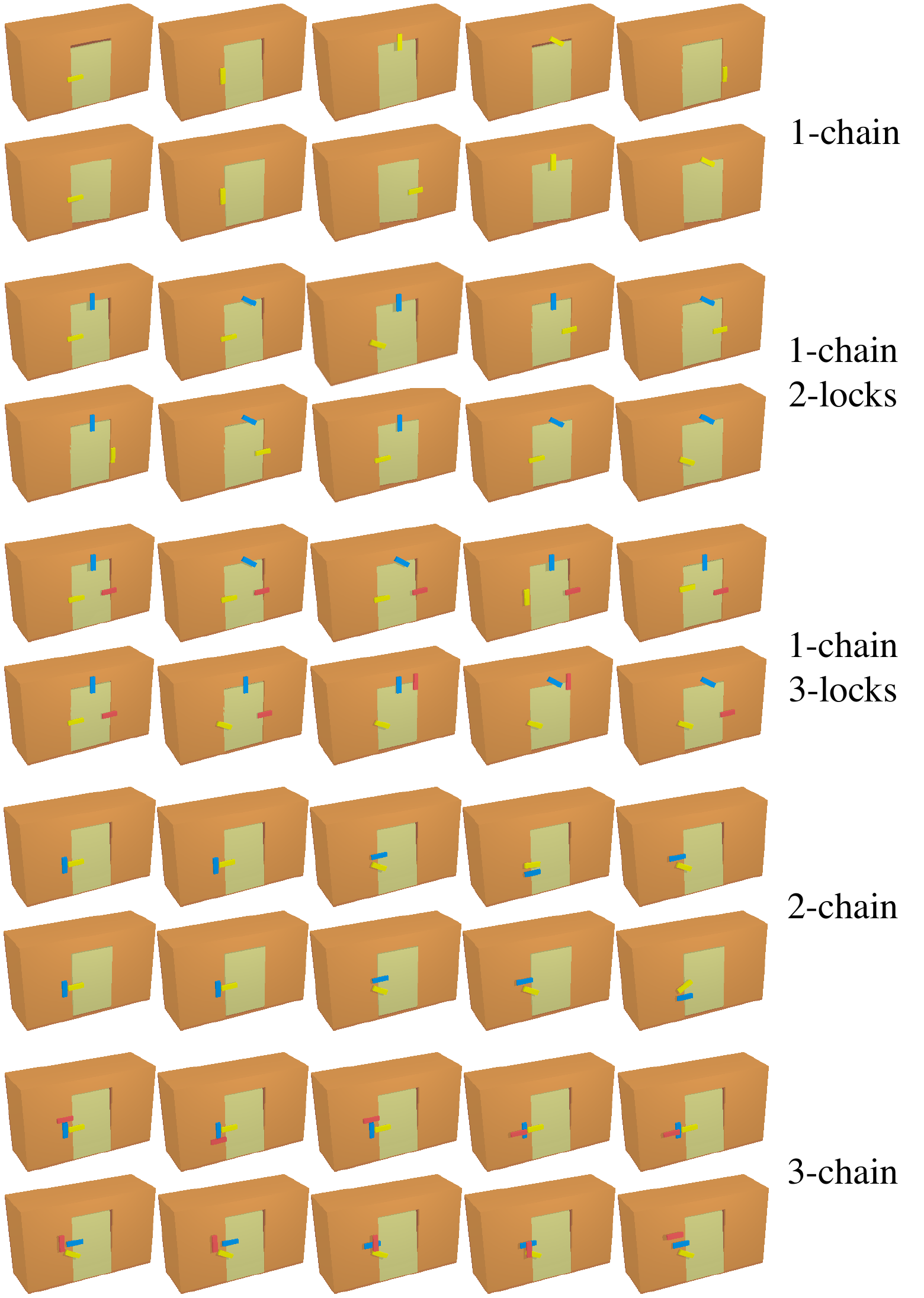}
    \caption{All instances from the PuzzleBoxes Benchmark.}
    \label{fig:puzzle_box_all}
\end{figure}

\noindent \textbf{PuzzleBox}
We aim to create a more comprehensive version of Thorndike's puzzle box experiment \cite{Thorndike1911-THOAI}, where a cat needs to escape a cage by exploring a locked door and finally opening it. We design the boxes with different level of difficulties by varying the number of locks and the number of decency. We show boxes with different levels in \fref{fig:puzzle_box_all}.

In the 1-chain N-locks setups, the door is blocked by a number of sliding locks and revolving locks. To solve the task, an agent needs to figure out which locks are blocking the door and how to manipulate these locks to resolve the blocking dependency. In these 1-chain setups, we assume the locks can be solved independently. To further test whether an agent can optimally explore by only researching on task decency locks, we also include boxes with dummy locks, e.g., locks that are already open and are not blocking the doors.

\begin{figure}
    \centering
    \includegraphics[width=\columnwidth]{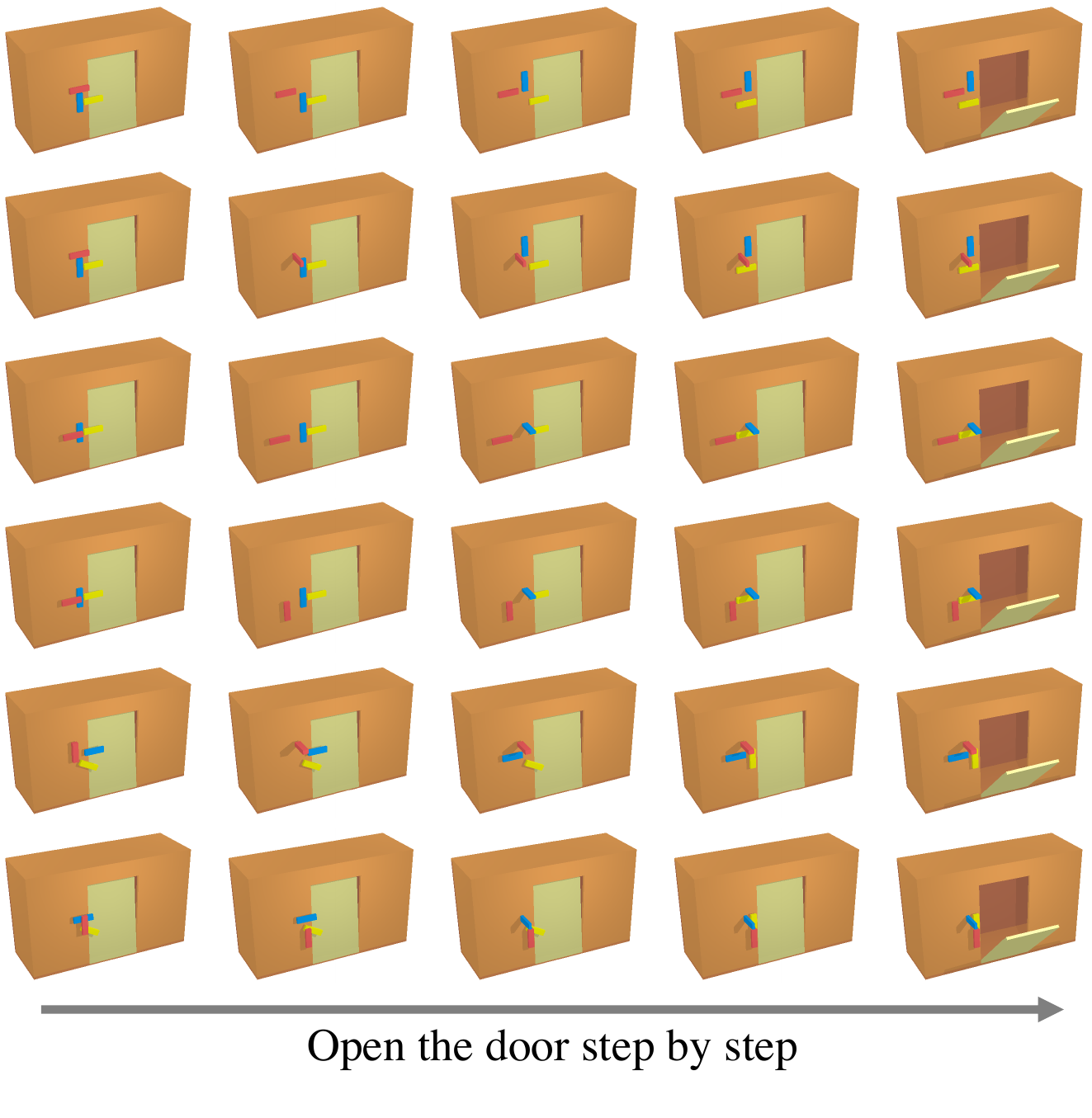}
    \caption{Visualizations of how the puzzle boxes will be opened by unlocking keys in 3-chain puzzle boxes.}
    \label{fig:puzzle_box_3chains}
\end{figure}

In the 2-chain and 3-chain setups, we test whether an agent can solve a chain of locks with dependency, i.e., the locks need to be operated in a specific order to fully unlock the door. In \fref{fig:puzzle_box_3chains}, we show some example design of the locks.
%tries to explore all existing locks without realizing which locks are blocking

\subsection{Implementation Details in the Joint Type Estimation Experiment}
\begin{figure}[t]
    \centering
    \includegraphics[width=\linewidth]{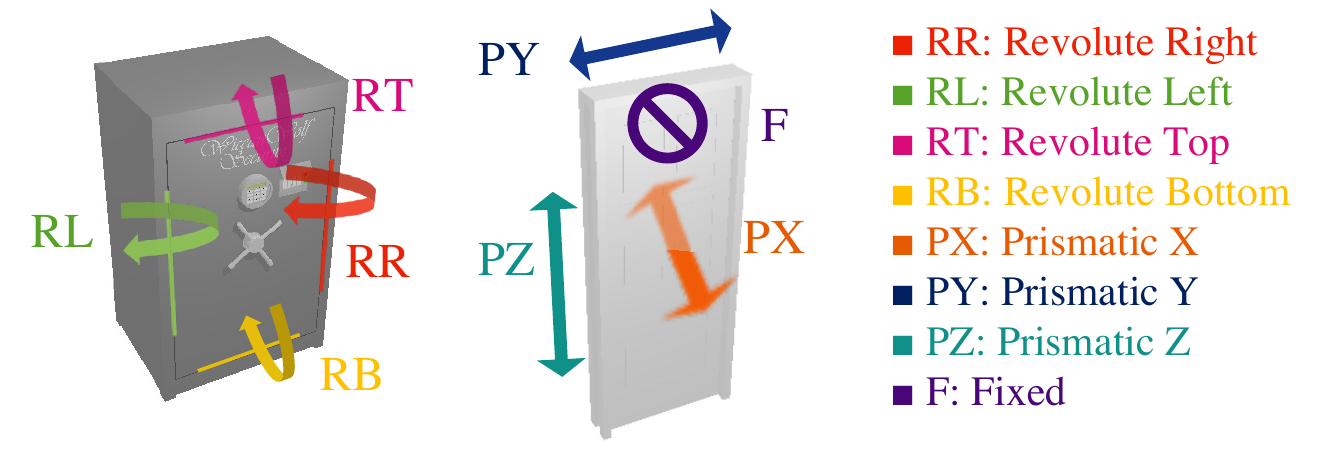}
    \caption{Joint configurations used for PartNet-Mobility dataset.}
    \label{fig:joints_config_sapien}
\end{figure}
\textbf{Types of Joints Considered}
To measure the performance, we cast the problem into an eight-way classification problem where the model classifies the target joint as one of the followings: (4) revolute joints attached to the right, left, top, or bottom of the 3D bounding boxes for the object part, (3) prismatic joints that moves along each of the X, Y, and Z axes, and a fixed joint. We illustrates the joints in \fref{fig:joints_config_sapien}.

%In Figure \todo{add figure}, we show some example design of the locks.
%\begin{itemize}
%    \item What kind of joint types we used (rev x 4, pris x 3, fixed x 1)
%    \item 
%\end{itemize}

\textbf{PN2}
Following~\cite{mo2021where}, we use PointNet++ network~\cite{qi2017pointnet++} and implementation~\cite{pytorchpointnet++} as a backbone feature extractor with four set abstraction layers with single-scale grouping for the encoder and four feature propagation layers for the decoder. We feed the extracted feature into a classification network which consists of two $256$-dim fully connected layers and 8-dim fully connected layer that classifies the input into the 8 possible joints.

\textbf{PN2+Ours} We can initialize the hypothesis distribution of the proposed model with prediction from PN2. However, 
we found that PN2 can generates almost zero weight on a correct hypothesis when it becomes over-confident about a false configuration. This can cause the proposed model to fail since all the particles are initialized with the over-weighted false configuration. To handle the problem, we impose a minimum weight $1/16$ on all the hypotheses to ensure all hypotheses are covered in the particle pool. We found the heuristic can significantly improve the overall performance (from $80.5 \%$ to $96.5\%$ on \textit{closed} setting with test categories).
%\item How to initialize prior distribution using PN2 estimation
%\begin{itemize}
%    \item How to initialize prior distribution using PN2 estimation
%\end{itemize}

%\subsection{Details in the Action Proposal and Affordance Map Experiment}
%\paragraph{Ours}
%\paragraph{Where2Act}

% For training the Where2Act network, we first randomly sample percentages for joint states for all movable links, open them, and get point cloud.

% For an applied action $a$ at point $p$, we estimate a likelihood $s_{a|p}$ for the success of the interaction parameterized by tuple $(p,a)\in SE(3)$. This network module $D_s$ is parameterized by an MLP with one hidden layer of size $128$. Given an input tuple $(f_p, a)$, we produce a scalar $s_{a|p}=D_s(f_p, a) \in [0,1]$.

% Furthermore, we also train a distance estimation model $D_d$ that estimates the relative distance caused by the action $d_{a|p}$ between the test object and the model's estimation.
% To train the model, we initialize the starting pose by uniformly sampling from $[0, 100]$ \%, where $0$ \% means fully-closed and $100$ \% fully-open.

%\subsection{Manipulation}
%\noindent \textbf{Where2Act} How to generate actions

%\section{Additional Algorithm Details for \modelname}

\subsubsection{Implementation Details in the RL Experiment for PuzzleBoxes} \label{subsec:rldetails}
This section provides more detailed implementation settings about the RL experiment described in \sref{subsec:puzzleboxes}.

\subsubsection{Environment}
\paragraph{State}
For a fair comparison to H-SAUR, which requires part segments and 3D information as point cloud, we define the state of the environment is two frames of RGBD image $I^{RGBD} \in \mathbb{R}^{100 \times 100 \times 8}$. The RGB image corresponds to part segments because all parts in PuzzleBoxes have unique color (see \fref{fig:puzzle_box_all}), and the depth image gives the 3D information to an RL agent.

\paragraph{Action}
The action of the agent consists of an interaction position and direction.
The position is defined by 2D continuous value $a^\text{pos} \in [-1, 1]^2$, and we find the closest pixel in the RGBD image and then find the corresponding 3D position to the pixel. This enables to make the action space smaller, and makes the RL agent easier to solve the task.
The interaction direction is defined by 3D continuous values $a^\text{dir} \in [-1, 1]^3$, and normalized so that it will be a unit vector.

\paragraph{Rewards}
Since training an RL agent entirely from a sparse reward is too hard to solve, we define a shaped reward to enhance exploration, which consists of changes in position of any joint $r^\text{shape} = \| \bm{\theta}_{t+1} - \bm{\theta}_{t} \|$, where $\bm{\theta}$ is a joint state vector that consists of all movable joints.

\paragraph{Terminal conditions}
An episode terminates with the same condition with other methods as described in \sref{subsec:puzzleboxes}: an agent opens the door outward for more than $60^\circ$, or the total step of an episode is over $100$.

\subsubsection{Agent}
We used CURL~\cite{laskin2020curl}, which uses contrastive learning to acquire a good image representation, for our RL experiments with the same hyper parameters used in the original paper.

\subsubsection{Training}
We used a random policy to collect $1$K transitions to a replay buffer before training an RL agent, and then trained the RL agent for $10$K timesteps. The configurations of PuzzleBoxes for train and test are different: we use $7$ configurations for training and $3$ configurations for testing, and we train RL agents for each PuzzleBoxes type separately.

% \clearpage
\section{Pseudo Code for \modelname}
Here we provide pseudo code for the proposed method in Algorithm \ref{alg:jntest}, \ref{alg:optimalact}, \ref{alg:solve puzzle box}, and \ref{alg:actdir}. Specifically, Algorithm  \ref{alg:solve puzzle box}, \ref{alg:actdir} handles joints with dependency, as described in Section 3.4. Since objects in the PartNet-Mobility Dataset do not have such dependency, they can be solved with Algorithm \ref{alg:jntest}, \ref{alg:optimalact} without the dependency check (highlighted in red in Algorithm 1). For objects in the PuzzleBox dataset, including the dependency check is critical to achieve reasonable performance.

%\ftnote{1. how to generate actions given direction,}

%\subsection{Goal-Orientated Manipulation}
\begin{algorithm}[tb]
\caption{Interactive Joint Type Estimation}
\label{alg:jntest}
\hspace*{\algorithmicindent} \textbf{Input} 
%Index of movable part $i \in \{1, ..., N_p \}$, 
Observed point cloud for the current movable part $O^{i}$, 
hypothetical joint configurations $\bm{ \theta} = \{ \theta^{(j)} | j \in \{1, 2, \cdots, J \} \}$,
%$\sigma(j) \in \{1, 2, \cdots, J\}$, 
number of particles $N^\text{particles}$,
threshold $\delta^\text{prob}$ to finish estimation, 
maximum steps to interact with the real-world object $N^\text{max}$ \\
\hspace*{\algorithmicindent} \textbf{Output} Estimated joint type $j^\ast$ and indexes of the collided movable parts $k$
\begin{algorithmic}[1]
\State Initialize a particle pool $\mathcal{S}^\text{state}= \emptyset$
\For{$k \gets 1$ to $N^\text{particles}$}
    \State Sample a joint configuration $\theta^{(\sigma(k))} \sim p_0(\bm{\theta})$ \Comment{Can use a learned prior distribution}
    \State Sample a joint upper limit $\theta^{\text{high}_k} \sim p_{\text{unif}[0, \theta_\text{MAX}]}$
    \State Sample a joint lower limit $\theta^{\text{low}_k} \sim p_{\text{unif}[-\theta_\text{MAX}, 0]}$
    %\State Sample a joint state $\theta^{\text{cur}_k} \sim p_{\text{unif}[\theta^{\text{low}_k}, \theta^{\text{high}_k}]}( \theta^{\text{cur}_k} )$
    \State Add the sampled particle $s^k = (\theta^{(\sigma(k))}, \theta^{\text{low}_k}, \theta^{\text{high}_k}, \theta^{\text{cur}_k} = 0)$ to the particle pool $\mathcal{S}^\text{state} \leftarrow \mathcal{S}^\text{state} \cup s^k$
\EndFor
\For{$t \gets 1$ to $N^\text{max}$}
    \State Infer an informative action $a^\ast$ using \aref{alg:optimalact}
    \State Apply the inferred action $a^\ast$ on the real-world object and observe the pointcloud $O_{t+1}$
    \State {\color{red}// Handling movable parts/joints with dependency}
 \If{{\color{red}Detect contacts between the current movable part and other movable parts}}
        \State Break the current loop
 \EndIf
        \ForAll {$k \gets 1$ to $N^\text{particles}$} 
            \State Apply the inferred action $a^\ast$ on $k$-th hypothetical object and observe the pointcloud $\hat{O}_{t+1}^k$
            \State Compute importance weight for the particle $w_k = \frac{1}{\text{dist} (O_{t+1}, \hat{O}^{k}_{t+1}) + \epsilon}$
        \EndFor
        \State Re-sample particles from $\mathcal{S}^\text{state}$ according to the importance weights. Compute updated posterior $p_{t+1}(\bm{\theta})$ from the particles. 
        %using \eref{eq:posterior} and $\{ \hat{O}_{t+1}^1, ..., \hat{O}_{t+1}^{N\text{particles}} \}$ and $O_{t+1}$
        \If {$\max_{j\in \{1,...,J \} } p_{t+1}(\theta^{(j)}) > \delta^\text{prob}$}
            \State Break the current loop
        \EndIf

\EndFor
\State \Return The most probable joint configuration $j^\ast = \argmax_{j\in \{1,...,J \}} p(\theta^{(j)})$ and {\color{red}indexes of the collided movable part $k$}
\end{algorithmic}
\end{algorithm}

% Algorithm to get an optimal action
\begin{algorithm}[tb]
\caption{Informative Action Selection}
\label{alg:optimalact}
%\la\State Reproduce a hypothetical object using the joint configuration $s$ in a physics engine
\hspace*{\algorithmicindent} \textbf{Input} Observed point cloud for the current movable part $O_i$, particle pool $\mathcal{S}^\text{state}$, number of particles for generating action $N^\text{particles-action}$, number of particle updates $N^\text{update}$ \\
\hspace*{\algorithmicindent} \textbf{Output} 
Informative action $a^\ast$
%The optimal action $a^\ast$ that maximally changes the joint state
\begin{algorithmic}[1]
\State Sample a joint configuration $s \sim \mathcal{S}^\text{state}$
\State Reproduce a hypothetical object using the joint configuration $s$ in a physics engine
%\ftnote{how to get the moving direction}

%\ftnote{termination criteria}
\State Initialize a particle pool $\mathcal{S}^\text{action}= \emptyset$
\For{$k \gets 1$ to $N^\text{particles-action}$}
    \State Sample a point to interact $p_k \sim O_i$
    \State Sample a force direction $r_k \sim \text{\{+x,-x, +y, -y, +z, -z\}}$
    \State Add the sampled particle $a_k = (p_k, r_k)$ to the particle pool $\mathcal{S}^\text{action} \leftarrow \mathcal{S}^\text{action} \cup a_k$
\EndFor
\For{$n \gets 1$ to $N^\text{update}$}
    % \State Initialize a likelihood pool $w = [0, ...] \in \mathbb{R}^{N^\text{particles-action}}$
    \For{$k \gets 1$ to $N^\text{particles-action}$}
        \State Apply the sampled action $a_k$ on the hypothetical object and observe the joint state $\theta_{t+1,k}^{s}$
        \State Compute importance weight for the particle $w_k = \| \theta_{t+1,k}^{s} - \theta_{t,k}^{s} \|$
    \EndFor 
    \State Re-sample particles from $\mathcal{S}^\text{action}$ according to the importance weights
    \State Add noise to the position $p_k \in \mathcal{S}^\text{action}$
\EndFor
\State \Return The most probable particle $a^\ast = a_{k^\ast}$, where $k^\ast = \argmax_{k \in \{ 1, ..., N^\text{particles-action}\} } w_k$
\end{algorithmic}
\end{algorithm}

% Algorithm to open the PuzzleBox's door
\begin{algorithm}[tb]
\caption{Unlock PuzzleBoxes}
\label{alg:solve puzzle box}
\hspace*{\algorithmicindent} \textbf{Input} Pointcloud of the target object $O,$ segmentation masks $m_1, m_2, \cdots, m_{N_v},$ target part ID $t$ to open, desired direction $d^t$ to displace the target part \\
\hspace*{\algorithmicindent} \textbf{Output} ``Success'' or ``Failure'' to solve the task
\begin{algorithmic}[1]
\State Obtain segmented point clouds $O_1, O_2, \cdots, O_{N_v}$ from $O$ and $m_1, m_2, \cdots, m_{N_v}$
\State Initialize the part-of-interest queue with the target part, e.g., the door, and its desired opening direction $q_\text{POI} = \{(O^t, d^t)\}$
\While{$|q_\text{POI}| \geq 0$}
    \State Get the most recently added object part and its moving direction $(O^l,d^l)$ from the queue $q_\text{POI}$
    \State Estimate the joint type of the target part $O^l$ using \aref{alg:jntest}
    \If {Detect collided parts during joint type estimation}
        \State Add the collided part $O^k$ and its unknown moving direction $d^k$ to the last of the part-of-interest list $q_\text{POI} \leftarrow q_\text{POI} \cup \{ (O^k, d^k) \}$
    \Else
        \If {The moving direction $d^l$ is ``unknown''}
            \State Compute the moving direction $d^l$ that resolves collision using \aref{alg:actdir}
        \EndIf
        \State Infer an optimal action $a^\ast$ on the estimated joint type using  \aref{alg:optimalact} while replacing the cost function $w_k = \| \theta_{t+1,k}^s - d^l \|$ in line.12 in \aref{alg:optimalact} and apply it on the real-world object
        \If {Goal is achieved (e.g., ``door is open'')}
            \State Terminate this experiment with ``Success''
        \ElsIf{Number of interactions exceeds $N^\text{max-int}$}
            \State Terminate this experiment with ``Failure''
        \ElsIf{part $O^{l}$ has been successfully displaced with $d^l$}
            \State Remove the current part and direction $q_\text{POI} \leftarrow q_\text{POI} \backslash \{ (O^l, d^l) \}$
        \EndIf
    \EndIf
\EndWhile
\State \Return ``Failure''
\end{algorithmic}
\end{algorithm}

\begin{algorithm}[tb]
\caption{Compute Action Direction to Resolve Collision}
\label{alg:actdir}
\hspace*{\algorithmicindent} \textbf{Input} Pointcloud of the current part of interest $O^l$, pointcloud of the collided object part $O^\text{c}$, estimated current part's joint configuration $s = \{ \theta^\text{low}, \theta^\text{high}, \theta^l \}$, number of samples to search for the desired joint state $N$\\
%, and the current joint state $\theta^s_t$ \\
\hspace*{\algorithmicindent} \textbf{Output} Desired joint state $\theta^\ast$ for the collided object part
\begin{algorithmic}[1]
\State Reproduce a hypothetical object using the joint configuration $s$ in a physics engine.
% \State Search a joint state $\theta^\ast (\theta^\text{low} \leq \theta^\ast \leq \theta^\text{high})$ that maximizes the pointcloud distance between $O^{l}_{\theta^\ast}$ and $O^\text{c}$, where $O^{l}_{\theta^\ast}$ is obtained by setting the joint configuration $s$ with joint position $\theta^\ast$ in a physics engine.
\State Obtain pointclouds $\{ O^{l}_{\theta_1}, ..., O^{l}_{\theta_N} \}$ of different joint states $\theta \in \{ \theta_1, ..., \theta_N \}$ by setting the joint configuration $s$ with joint position $\theta$ in a physics engine, where $\theta$ is evenly spaced between $\theta^\text{low}$ and $\theta^\text{high}$.
\State \Return The joint state 
% $\theta^\ast$ that maximizes the point cloud distance between the current part and the collided part 
$\theta^\ast = \argmax_{\theta \in \{ \theta_1, ..., \theta_N \}} \text{dist}(O^{l}_{\theta}$, $O^\text{c})$
% \State \Return The inferred joint state $d_l = \theta^\ast$
\end{algorithmic}
\end{algorithm}

\end{document}